# Event knowledge in large language models: the gap between the impossible and the unlikely

Carina Kauf*[,1,2], Anna A. Ivanova*[,1,2], Giulia Rambelli[3], Emmanuele Chersoni[4], Jingyuan Selena She[1,2], Zawad Chowdhury[5], Evelina Fedorenko[1,2], Alessandro Lenci[6]

[1]*Department of Brain and Cognitive Sciences, Massachusetts Institute of Technology*
[2]*McGovern Institute for Brain Research*
[3]*Department of Modern Languages, Literatures and Cultures, University of Bologna*
[4]*Department of Chinese and Bilingual Studies, Hong Kong Polytechnic University*
[5]*Department of Mathematics, University of Washington*
[6]*Department of Philology, Literature, and Linguistics, University of Pisa*

** The two lead authors contributed equally to this work.*

**Corresponding authors**: Carina Kauf (ckauf@mit.edu) and Anna Ivanova (annaiv@mit.edu)

# Abstract

Word co-occurrence patterns in language corpora contain a surprising amount of conceptual knowledge. Large language models (LLMs), trained to predict words in context, leverage these patterns to achieve impressive performance on diverse semantic tasks requiring world knowledge. An important but understudied question about LLMs' semantic abilities is whether they acquire generalized knowledge of common events. Here, we test whether five pre-trained LLMs (from 2018's BERT to 2023's MPT) assign higher likelihood to plausible descriptions of agent-patient interactions than to minimally different implausible versions of the same event. Using three curated sets of minimal sentence pairs (total n=1,215), we found that pre-trained LLMs possess substantial event knowledge, outperforming other distributional language models. In particular, they almost always assign higher likelihood to possible vs. impossible events (*The teacher bought the laptop* vs. *The laptop bought the teacher*). However, LLMs show less consistent preferences for likely vs. unlikely events (*The nanny tutored the boy* vs. *The boy tutored the nanny*). In follow-up analyses, we show that (i) LLM scores are driven by both plausibility and surface-level sentence features, (ii) LLM scores generalize well across syntactic variants (active vs. passive constructions) but less well across semantic variants (synonymous sentences), (iii) some LLM errors mirror human judgment ambiguity, and (iv) sentence plausibility serves as an organizing dimension in internal LLM representations. Overall, our results show that important aspects of event knowledge naturally emerge from distributional linguistic patterns, but also highlight a gap between representations of possible/impossible and likely/unlikely events.

# 1. Introduction

## 1.1 Language and event knowledge

A vital component of human intelligence is our ability to learn, store, and flexibly use rich, structured knowledge about the world. World knowledge spans different domains (from physical properties to social conventions) and covers different types of information, including knowledge of objects, agents, actions, and ideas. One important component of world knowledge is our *generalized event knowledge (GEK)* – templates of common events observed in the world (e.g., McRae & Matsuki, 2009). Humans acquire GEK both through sensorimotor experiences (i.e., from participating in and observing events in the world) and through linguistic experiences (i.e., from event descriptions generated by other people) (Dove, 2020; Günther et al., 2020; Dove, 2023). Here, we ask: To which extent can GEK be learned simply by tracking distributional properties of linguistic input?

On the one hand, positing that GEK can be learned from language alone appears to contradict the fact that in humans, much of conceptual knowledge is innate (e.g., Spelke & Kinzler, 2007) or learned through direct experience (Meteyard & Vigliocco, 2008). On the other hand, co-occurrence patterns learned from language exhibit a remarkable degree of correspondence with distributional spaces learned through other modalities, like vision (Lewis et al., 2019; Roads & Love, 2020; Abdou et al., 2021; Patel & Pavlick, 2021; Sorscher et al., 2021).This alignment suggests that language-based distributional information might be able to replace other modalities as a source of world knowledge (Louwerse, 2011). Indeed, knowledge of events is abundantly represented in language corpora, presumably because humans typically communicate events that are, were, or will be happening in the world (e.g., McRae & Matsuki, 2009). Consequently, the GEK that can be learned from distributional linguistic knowledge might faithfully reflect the GEK that people typically acquire multimodally.

## 1.2 LLMs as models of semantic knowledge

To disentangle contributions of distributional linguistic knowledge from other sources of information — a feat that's difficult to accomplish in humans (e.g., Kim et al., 2019; Lewis et al., 2019; Ostarek et al., 2019) — we turn to large language models (LLMs). LLMs are the latest generation of distributional semantic models (Lenci & Sahlgren, 2023), which learn rich semantic representations through tracking word co-occurrence patterns in text in service of their training objective, i.e., predicting the next/a missing word from a given linguistic context. A wealth of research has demonstrated that distributional semantic models can explain a broad range of phenomena in human cognition, e.g., synonym judgment (Landauer & Dumais, 1997; Levy et al., 2017), similarity judgements (Hill et al., 2015), semantic priming effects in word naming and lexical decision tasks (Mandera et al., 2017). This makes them a useful tool for probing language representations in the human mind and for understanding what kind of information can be learned from text alone.

We focus on LLMs that have been trained on large, general text corpora with a word-in-context prediction objective, often referred to as “pre-trained” LLMs. The word-prediction objective enables these models to learn rich amounts of knowledge without being constrained by specific task demands; moreover, this objective closely parallels the next-word-prediction behavior observed in humans (e.g., McRae et al., 1998; Altmann & Kamide, 1999; Traxler et al., 2002; Levy, 2008; Kutas & Federmeier, 2011; Mani & Huettig, 2012; Smith & Levy, 2013; Kuperberg & Jaeger, 2016; Shain, Blank et al., 2020), making it a cognitively plausible training function for distributional language models (e.g., Schrimpf et al., 2021; Goldstein et al., 2022; Hosseini et al., 2022). Due to the focus on models that capture the task-agnostic distributional language spaces, fine-tuned LLMs are beyond the scope of this paper.

LLMs today generate grammatically correct, syntactically varied, and semantically relevant texts, indicating that these models have essentially mastered *formal* linguistic competence, i.e., knowledge of the rules and patterns that govern natural language (Mahowald, Ivanova et al., 2023; Contreras Kallens et al., 2023; see also Piantadosi, 2023). However, their *functional* linguistic competence, i.e., their general knowledge and reasoning skills as expressed through language, remain highly debated (e.g., Bender & Koller, 2020; Marcus, 2020; Mahowald et al., 2023). Despite their seemingly remarkable success across a variety of tasks, such as generating syntactically and semantically coherent paragraphs of text (Brown et al., 2020), sentiment analysis and logical inference (e.g., Devlin et al., 2018; Liu et al., 2019; Radford et al., 2019; Yang et al., 2019), closed-book question answering (QA) (Roberts et al., 2020), theory of mind (Kosinski, 2023; Shapira et al., 2023; Trott et al., 2023), and certain aspects of commonsense reasoning (e.g., Zellers et al., 2018), a closer examination of LLM performance reveals that they frequently rely on low-level word-co-occurrence patterns, which, when removed, drastically affect LLM performance (e.g., She et al., 2023; Ullman, 2023). This performance pattern stands in contrast to human performance, which is typically robust to such low-level variations (although see Dasgupta et al., 2022; Lampinen, 2022 for calls to not overestimate human performance).

Studies of world knowledge in LLMs have likewise produced mixed results. On the one hand, even non-finetuned LLMs perform well on multiple tasks designed to probe world knowledge, such as the Winograd Schema Challenge (WSC; Levesque et al., 2012), the Story Cloze Test (SWAG; Zellers et al., 2018), and the Choice of Plausible Alternatives Test (COPA; Roemmele et al., 2011), so much so that some authors have proposed and evaluated their use as off-the-shelf knowledge base models (Petroni et al., 2019; Roberts et al., 2020; Tamborrino et al., 2020; Kassner et al., 2021). On the other hand, studies using more fine-grained tests have shown that world knowledge in contemporary LLMs is often brittle and depends strongly on the specific way the problem is stated (McCoy et al., 2019; Niven & Kao, 2019; Ettinger, 2020; Kassner & Schütze, 2020; Ravichander et al., 2020; Elazar et al., 2021a; 2021b; Pedinotti et al., 2021; Ribeiro et al., 2020). For example, some authors have noted that, when low-level co-occurrence statistics are properly controlled for, LLMs that were considered to have high accuracy on world knowledge tasks start to perform randomly (Elazar et al., 2021b; Sakaguchi et al., 2021), highlighting the potential discrepancy between the word-in-context prediction objective (which benefits from tracking surface-level statistics) and world knowledge acquisition (which should be invariant to surface-level statistics).

## 1.3 LLMs as models of GEK

In principle, LLMs should be well-posed to acquire GEK. First, significant subparts of GEK are readily available in language co-occurrence statistics. This is evidenced by the success of relatively small distributional semantic models, such as distributional selectional preference models (Padó et al., 2006; Erk, 2007; Padó et al., 2007), or a more recent Structured Distributional Model (Chersoni et al., 2019), which explicitly represents GEK as a distributional event graph of syntagmatic relations extracted from dependency-parsed corpora (see also Sayeed et al., 2015; Santus et al., 2017) on thematic fit modeling tasks (Vassallo et al., 2018). Furthermore, Elman & McRae (2019) show that a small recurrent network trained with a string prediction objective is able to extract GEK about certain events from a small set of curated training set sentences.

Second, the increased scale of LLMs in comparison to earlier generations of distributional semantic models of GEK (such as SDM) — for example in terms of their numbers of parameters, training data size, or context window size — should be conducive for learning much richer patterns of event knowledge than traditional distributional methods (e.g., Erk, 2012). They should therefore be able to generalize their representations more easily to unseen event descriptions. Moreover, the size of recent LLMs allows for the verbatim memorization of a large number of text sequences (e.g., Carlini et al., 2021, 2022), which necessarily contain event descriptions.

Third, the word-in-context prediction objective that LLMs are trained with is closely tied to GEK in humans. A range of psycholinguistic studies shows that humans continuously predict upcoming words in service of efficient and resource-optimal language comprehension (e.g., McRae et al., 1998; Altmann & Kamide, 1999; Traxler et al., 2002; Levy, 2008; Kutas & Federmeier, 2011; Mani & Huettig, 2012; Smith & Levy, 2013; Kuperberg & Jaeger, 2016; Shain, Blank et al., 2020) and that, in doing so, they rely extensively on their GEK to dynamically update their expectations (Ferretti et al., 2001; McRae et al., 2005; Hare et al., 2009; McRae & Matsuki, 2009; Bicknell et al., 2010; Matsuki et al., 2011). Knowledge of events is helpful for predicting words from context since it helps restrict the range of possible, plausible continuations to those that are compatible with the event interpretation triggered by the linguistic context. For example, in the context *Donna used the shampoo to wash her filthy* ___, the integration of the lexical items *wash* and *shampoo* triggers a washing sub-event that renders the mention of *hair* unsurprising relative to other possible patients of a generalized washing event, such as *car* (Matsuki et al., 2011). Given the utility of deploying GEK for predicting words from context in humans, a possible strategy for LLMs to succeed in word-in-context prediction is to likewise construct rich internal, generalizable representations of event knowledge from distributional linguistic information.

Nonetheless, the complexity and scale of GEK makes modeling it a challenging target for any model relying on distributional semantic knowledge. First, language is sparse and the possible combinatorial space of events and their arguments is vast, even in the relatively small domain of transitive agent-patient interaction events that we focus on here. To assess the plausibility of an arbitrary event, a successful model of GEK must therefore acquire robust, generalizable representations of a vast number of actions and their associated restrictions on event participants. Many traditional and current distributional models have been argued to lack the representations of these building blocks for more complex semantic structures (Zhu et al., 2018; Pedinotti et al., 2021; Lenci, 2023; Lenci & Sahlgren, 2023). The acquisition of GEK is complicated even more because the frequency with which events are reported in the pragmatically influenced texts available in the world, is not a robust indicator of the frequency with which they occur in the real world (Gordon & Van Durme, 2013; see also **Section 4.3**). Thus, it remains unclear whether the latest generation of distributional semantic models acquire human-like robust, generalizable GEK from text co-occurrence statistics.

## 1.4 This study

In this work, we test whether pre-trained LLMs encode human-like generalized world knowledge in the domain of events. The term "event" has different meanings across disciplines and can encompass both an individual action or a sequence of several actions (Zacks et al., 2007; Zacks, 2020; see Kuperberg, 2021 for discussion). Some research on event knowledge in LLMs, for example, asks whether LLMs trained on word-in-context prediction encode human-like knowledge of event boundaries, investigating their capacity to replicate a fundamental aspect of human cognitive processing related to understanding sequential events in narratives (Kumar et al., 2022; Wang et al., 2022; Michelmann et al., 2023).

Here, we define an event in the linguistic tradition, as a singular action along with the entities that participate in that action in a particular role (e.g., Fillmore, 1967; Jackendoff, 1987; Dowty, 1989). We focus on transitive two-participant events: agent-patient interactions, such as *The teacher bought the laptop*. Our goal here is to explore implicit knowledge of events in LLMs, operationalized as a systematic preference for generating descriptions of plausible over implausible events. We investigate five open-source LLMs: MPT, GPT-J, GPT-2, RoBERTa, and BERT, as well as a range of non-LLM distributional models.

We hypothesize that, if GEK emerges naturally from the word-in-context prediction objective, pre-trained LLMs should treat plausible sentences as more likely than implausible sentences. If, on the other hand, distributional knowledge in pre-trained LLMs does not consistently reflect event knowledge, their event representations would fail to systematically align with GEK.

To minimize the effect of confounding factors, we use highly controlled, syntactically simple minimal sentence pairs drawn from three datasets. In two datasets (Datasets 1 and 3), plausibility is manipulated via swapping the agent and patient of the sentence (e.g., *The teacher bought the laptop* vs *The laptop bought the teacher*). This manipulation ensures identical word-level content within a sentence pair, such that the plausibility inference requires identifying the role played by each participant (e.g., *teacher* = agent, *laptop* = patient). In Dataset 2, plausibility is manipulated by replacing the patient of the event (e.g., *The actor won the award/battle*). The three datasets were selected to span event descriptions across a range of event participant compositions (interactions between two animate or one animate and one inanimate event participant) as well as varying degrees of semantic incongruence of the manipulated sentence (ranging from impossible to moderately implausible events). We focus on our largest dataset (Dataset 1, see **Methods**) for most analyses but show in **SI** that the findings extend to other datasets. We restrict ourselves to simple event descriptions in English, with the caveat that our results might not generalize to other languages (Atari et al., 2023; Blasi et al., 2022).

In **Sections 3.1** and **3.2**, we ask whether LLMs and humans assign higher likelihood scores to descriptions of plausible events compared to their implausible counterparts. In **Sections 3.3** and **3.4**, we investigate the degree to which these scores are *generalized*, i.e., abstracted away from the surface-level properties of the input. Finally, we conduct detailed analyses of LLM performance by studying their error patterns (**Section 3.5**) and the nature of their internal representations of event plausibility (**Section 3.6**).

To foreshadow our key results, we find that LLMs possess substantial implicit event knowledge and outperform strong baseline models. In particular, they systematically prefer events that are possible (e.g., *The teacher bought the laptop*) to events that are, in the absence of contextual information, impossible (e.g., *The laptop bought the teacher*). However, LLMs are less consistent when distinguishing events that are likely (e.g., *The nanny tutored the boy*) from events that are unlikely but not impossible (e.g., *The boy tutored the nanny*), although their performance is still significantly above chance. Thus, we conclude that possible and impossible events naturally segregate in the distributional linguistic space, whereas likely and unlikely events segregate to a lesser extent, suggesting that *some but not all kinds* of event knowledge can be naturally learned by tracking distributional linguistic knowledge.

# 2. Methods

## 2.1 Sentence sets

We compare event plausibility scores in humans and language models using three sentence sets adapted from previous cognitive science and neuroscience studies (see **Tables 1** and **2** for a summary):

***Dataset 1 - main*** *(based on Fedorenko et al., 2020)***.** This sentence set contains 391 items, each of which includes **(i)** a plausible active sentence that describes a transitive event in the past tense (e.g., *The teacher bought the laptop)* and **(ii)** the implausible version of the same sentence, constructed by swapping the noun phrases (NPs) (*The laptop bought the teacher)*. The dataset also includes passive voice versions of the same sentences (*The laptop was bought by the teacher* and *The teacher was bought by the laptop).* Further, 249 of the 391 items are grouped into pairs where the sentences consist of words with synonymous, or closely related, meanings (e.g., *The teacher bought the laptop* and *The instructor purchased the computer*). For simplicity, we call those sentences “synonymous” throughout the paper.

The items are split into two types: (1) animate-inanimate (AI) items (e.g., *The teacher bought the laptop* vs. *The laptop bought the teacher;* n=128; 76 with synonyms); (2) animate-animate (AA) items (e.g., *The nanny tutored the boy* vs. *The boy tutored the nanny;* n=129; 82 with synonyms). Due to the animacy differences, the role reversal manipulation on AI sentences often violates the animacy selectional restrictions on the verb, making the sentence mostly semantically impossible, whereas the plausibility violations in AA sentences are more graded. Finally, the dataset includes a set of animate-animate, reversible (AA-control) items (n=134; 78 with synonyms), where both event participants are animate and both agent-patient combinations are plausible (e.g., *The cheerleader kissed the quarterback* vs. *The quarterback kissed the cheerleader*) and that we used as control in some of the analyses.

***Dataset 2 (DTFit;** based on Vassallo et al., 2018).* This sentence set contains 395 items, each of which includes **(i)** a plausible active sentence that describes a transitive event in the past tense, where the animate agent entity is interacting with an inanimate patient entity that is prototypical/canonical for the agent (e.g., *The actor won the award),* and **(ii)** the less plausible version of the same sentence, constructed by varying the inanimate patient entity (*The actor won the battle)*. Plausibility depends on the entire <agent, verb, patient> triple rather than just on the <agent, verb> or <verb, patient> combination. All sentence pairs in this dataset describe interactions between an animate agent and an inanimate patient, making them most comparable to the AI sentence pairs from Dataset 1. However, unlike in Dataset 1, word content and not word order distinguishes between plausible and implausible sentences within a pair. Note further that the plausibility manipulation in this sentence set is graded: the events can be described as typical/atypical rather than possible/impossible.

***Dataset 3** (based on Ivanova et al., 2021)***.** This sentence set contains 38 items, each of which includes **(i)** a plausible active sentence that describes a transitive event in the present tense (e.g., *The cop is arresting the criminal),* and **(ii)** the implausible version of the same sentence, constructed by swapping the NPs (*The criminal is arresting the cop)*. All sentence pairs in this dataset describe non-reversible interactions between two animate entities, making them comparable to the AA sentence pairs from Dataset 1. As in Dataset 1, only word order but not word content distinguishes between plausible and implausible sentences within a pair.

The majority of the sentences in Datasets 1 and 3 and all sentences in Dataset 2 use single nouns as subjects and objects; a small subset of sentences in Datasets 1 and 3 uses multi-word noun phrases (e.g., social worker). All active voice sentences in Datasets 1 and 2 and most sentences in Dataset 3 use the structure “Subject-Verb-Direct Object”; a small subset of sentences in Dataset 3 also contain indirect objects (*A doctor is using a stethoscope on the patient*). All datasets can be found on https://github.com/carina-kauf/lm-event-knowledge.

***Table 1***. *Sentence manipulations in Dataset 1.*

| Item Type | Plausible? | Possible? | Sentence |
|---|---|---|---|
| animate-inanimate (AI) | yes | yes | The teacher bought the laptop. |
| | no | no | The laptop bought the teacher. |
| animate-animate (AA) | yes | yes | The nanny tutored the boy. |
| | no | yes | The boy tutored the nanny. |

***Table 2***. *Sentence manipulations across the three datasets.*

| Sentence Set | Plausible? | Voice | Synonym # | Sentence |
|---|---|---|---|---|
| **Dataset 1** *(Fedorenko et al., 2020)* | yes | active | 1 | The teacher bought the laptop. |
| | | | 2 | The instructor purchased the computer. |
| | | passive | 1 | The laptop was bought by the teacher. |
| | | | 2 | The computer was purchased by the instructor. |
| | no | active | 1 | The laptop bought the teacher. |
| | | | 2 | The computer purchased the instructor. |
| | | passive | 1 | The teacher was bought by the laptop. |
| | | | 2 | The instructor was purchased by the computer. |
| **Dataset 2** *(Vassallo et al., 2018)* | yes | active | - | The actor won the award. |
| | no | active | - | The actor won the battle. |
| **Dataset 3** *(Ivanova et al., 2021)* | yes | active | - | The cop is arresting the criminal. |
| | no | active | - | The criminal is arresting the cop. |

## 2.2 Human data collection

For all three sentence sets, we compared language model predictions with human plausibility judgments. Human judgments for Dataset 2 had been previously collected by Vassallo et al., (2018) on Prolific, a web-based platform for collecting behavioral data. Participants in this experiment answered questions of the form "*How common is it for an actor to win an award*?" on a Likert scale from 1 (very atypical) to 7 (very typical). Human judgments for Dataset 1 and 3 were collected on Amazon Mechanical Turk, another web-based platform. Here, participants evaluated the extent to which each sentence was "plausible, i.e., likely to occur in the real world" on a Likert scale from 1 (completely implausible) to 7 (completely plausible). The protocol for the study was approved by MIT's Committee on the Use of Humans as Experimental Subjects (COUHES). All participants gave written informed consent in accordance with protocol requirements.

For Dataset 1 (our main dataset), we recruited 966 participants, restricting our task to participants with IP addresses in the US. The sentences were divided into 32 experimental lists such that each of the items occurred only in one of its versions in any given list. The median response time was 20.6 min. Each participant completed between 1 and 3 lists (mean=1.1).

Participants were included in the analyses if they satisfied all the following criteria: i) self-reported location ("USA"), ii) native English proficiency (evaluated via self-report and two sentence completion trials), iii) fewer than 20% of blank responses, and iv) accurate responses to attention checks ("Please select the leftmost/rightmost option"). We additionally filtered participants based on their responses to the AI items (*The teacher bought the laptop* vs. *The laptop bought the teacher*), retaining participants with a minimum plausibility difference of 1 point (out of 7) between plausible and implausible items in this condition. These criteria left data from 658 participants for analysis. Each sentence had a minimum of 18 ratings (average: 22.9 ratings; maximum: 27 ratings). Participants were paid $4.25 (estimated completion time was 25 min), with payment contingent only on the attention-check questions and excessive blank responses (>30%).

For Dataset 3, we recruited 100 participants, restricting our task to participants with IP addresses in the US. The sentences were divided into 2 experimental lists and each of the items occurred only in one of its versions in any given list. The median response time was 15.7 min. Each participant completed 1 list. We filtered the data using the same criteria as for Dataset 1, except for the sentence completion trials for assessing English proficiency (which were not included) and the minimum plausibility difference criterion. The inclusion/exclusion criteria left data from 96 participants for analysis (48 ratings per sentence). Participants were paid $2.70, with payment contingent only on the attention-check questions and excessive blank responses (>30%).

## 2.3 Model description and score estimation

### 2.3.1 Large language models (LLMs)

We tested five attention-based Transformer (Vaswani et al., 2017) language models: MPT (The MosaicML NLP Team, 2023), GPT-J (Wang & Komatsuzaki, 2021), GPT-2 (Radford et al., 2019), RoBERTa (Liu et al., 2019), and BERT (Devlin et al., 2018). GPT-2, GPT-J, and MPT are unidirectional (aka autoregressive or causal) models, trained to predict upcoming words based only on left context (e.g., *The teacher bought the <MASK>*). BERT and RoBERTa are bidirectional models; their primary training task is predicting masked words in the input based both on left and right context (e.g., *The <MASK> bought the laptop*). For all Transformer models, we used pre-trained implementations available via the HuggingFace Transformers library (Wolf et al., 2020). Specifically, we investigated the following model instantiations: *mpt-30b* (Number of layers, L=48; Hidden size, H=4096), *gpt-j-6B* (L=28, H=4096), *gpt2-xl* (L=28, H=4096), *roberta-large* (L=24, H=1024), *bert-large-cased* (L=24, H=1024), i.e., the largest pre-trained version per model available via HuggingFace. See **Table S1** for more information about the LLMs' architecture and training.

For the unidirectional LLMs, we define the sentence score as the sum of the log-probabilities of each token $w_i$ in the sequence, conditioned on the preceding sentence tokens $w_{<i}$.

For the bidirectional LLMs, we define the sentence score as a variant of the sentence's pseudo-log-likelihood score (PLL). The original PLL scoring method defines a sentence's score as the sum of the log-probabilities of each token given all other tokens (Wang & Cho, 2019; Salazar et al., 2020). This method, however, yields inflated scores for multitoken words (Kauf & Ivanova, 2023). Here, we use the improved the PLL scoring method introduced by Kauf & Ivanova (2023), which avoids this bias by masking not only the target token, but also all within-word tokens to the right of the target during inference. We show in **Figures S11** and **S12** that sentence generation likelihood is a more robust indicator of event knowledge in bidirectional LLMs than other prediction-based metrics, such as last-word prediction probability or verb prediction probability for our datasets.

To encourage transparency in the NLP community, we do not report results from closed models, such as GPT-3. We also do not report results from models that have been fine-tuned on additional objectives, such as reinforcement learning from human feedback: our goal is to specifically test world knowledge encoded in the distributional patterns learned via word-in-context prediction.

### 2.3.2 Baseline models

To investigate whether knowledge of event plausibility depends on specific linguistic patterns, we additionally compared the performance of the LLMs against four baseline models. This comparison allows us to evaluate the added value of LLMs in comparison to more "traditional" but less complex distributional semantics models, typically trained on a much smaller amount of data (Lenci & Sahlgren, 2023).

**TinyLSTM** is a two-layer LSTM recurrent neural network trained with a next-word prediction objective on the string data from the 1-million-word English Penn Treebank §2-21 (Marcus et al., 1993). Like for unidirectional LLMs, a sentence score for TinyLSTM is estimated as the sum of negative log probabilities of each token conditioned on the preceding tokens. The model is available through the LM Zoo library (Gauthier et al., 2020).

**Thematic fit** models the degree of semantic compatibility between an event's "prototype" verb argument, calculated from distributional text information (McRae et al., 1998), and the role filler proposed by the sentence. We follow the approach for calculating prototypical argument representations by Lenci (2011) and compute a prototype representation for the event patient slot as the centroid vector representations from the most associated entities with the predicate and agent in the sentence. However, instead of computing updates to the prototype using Distributional Memory vectors (as in Lenci, 2011), we here do the same computations using FastText (Bojanowski et al., 2017) static embeddings (see also Rambelli et al., 2020). A sentence's plausibility score is computed as the cosine similarity between the FastText embedding of the proposed patient and the relevant prototype vector.

The **Structured Distributional Model** (SDM; Chersoni et al., 2019) is a model of thematic fit that computes both a *context-independent* and a *context-dependent* representation of the prototype role filler based on the current linguistic context. The context-independent representation is obtained via summing the FastText embeddings of all lexical items in the current linguistic context. The context-dependent representation is derived based on a dynamic representation of the context: given the lexical items in the current context and the syntactic function of the next word to be predicted, SDM queries a distributional event graph (DEG) to retrieve the words with the strongest statistical associations with those items for the target function (the DEG was extracted from a large number of dependency-parsed corpora: words are linked with their syntactic collocates and the links weighted with mutual information scores). It then computes the centroid of the FastText embeddings associated with the highest-ranked lexical entities according to DEG. Finally, a sentence's plausibility score is calculated as the sum of the SDM thematic fit scores for each verb argument (in our case: agent and patient), whereby each score is derived as the average cosine similarity of the argument filler's representation with the context-dependent and context-independent prototype representations of the role.

Lastly, the **PPMI-syntax** model quantifies the statistical association between verbs and their dependents (marked for syntactic role, i.e., $PPMI(arrest, cop_{subj}) \neq PPMI(arrest, cop_{obj})$) in terms of Positive Pointwise Mutual Information (PPMI). It is trained on the same dependency-parsed corpus as SDM. We apply Laplace smoothing and compute the plausibility score of a sentence as the PPMI score between the verb and the subject plus the PPMI score between the verb and the object.

See **SI 2** for additional baseline model description details.

## 2.4. Binary accuracy estimation

To assess GEK in language models and in humans, we present them with minimally different plausible vs. implausible event descriptions (**Section 2.1**). We evaluate their ability to assign a higher score to the plausible event description than the implausible one (**Sections 2.2** and **2.3**). Human scores were averaged to obtain a single score for each sentence. For each sentence pair, we assigned a score of 1 if the model/human subject pool succeeded on this task, i.e., if a higher score was assigned to the plausible version of the sentence and 0 otherwise.

## 2.5. Word frequency estimation

To account for potential effects of word frequency, we estimated the average frequency of the word/phrase denoting the agent, patient, and verb of each sentence, as well as the average frequency of all words in the sentences. Frequency was operationalized as the log of the number of occurrences of the word/phrase in the 2012 Google NGram corpus. Laplace smoothing was applied prior to taking the log.

## 2.6. Probing analysis

To investigate the emergence of explicit plausibility information in LLMs, we trained a decoding probe to distinguish plausible and implausible sentences from their embeddings at different LLM layers. Separate logistic regression classifiers were trained for each model layer and the static word embedding space of the models. For each sentence, the input was the model-specific sequence summary token; the output was a binary plausibility label. The choice of model-specific sequence summary tokens followed the default settings from Huggingface Transformers: for the bidirectional LLMs, BERT and RoBERTa, we used the representation of the special token [CLS], which was prepended to each stimulus and was designed and trained specifically for sequence classification tasks. For the unidirectional LLMs, GPT-J and GPT-2, we prepared the stimulus by adding the [EOS] token to the beginning and end of the sequence and used the representation of the final token as the sequence's summary representation. For all analyses, probes were trained using 10-fold cross-validation, ensuring that plausible and implausible versions of the same sentence remain in the same split (train or test). To estimate the best-case model performance, we computed empirical ceiling values by training probes on the average human plausibility ratings for each sentence. The probe setup and the cross-validation procedure for ceiling probes were the same as for LLM probes.

To probe the generalization ability of the LLMs, we trained the classifiers on just one type of sentence (either on specific animacy combinations, AI or AA, or specific voice, active or passive) and evaluated the performance on the held-out type.

We used sklearn's (Pedregosa et al., 2011) Logistic Regression module with a liblinear solver for all probing analyses.

## 2.7. Statistical analyses

**Binary accuracy**. Binary accuracy results were compared to chance performance of 0.5 using a binomial test. Tests of equal proportion were used to compare model performance to human performance, as well as AI sentence accuracy to AA sentence accuracy within each metric.

**Correlations**. All reported correlations are Pearson correlations. Correlation significance was assessed using the test for correlation for paired samples (cor.test in R). Model correlation was compared to human correlation using the *cocor* package's (Diedenhofen & Musch, 2015) implementation of (Raghunathan et al., 1996) test for nonoverlapping correlations based on dependent groups.

**Mixed effects modeling**. We fitted separate linear mixed effects models to human ratings and each language model's scores. The key predictors for Dataset 1 were plausibility, item type (AI vs. AA vs. AA-control), and voice (active vs. passive), as well as interactions between them. We also included agent, patient, verb, and average sentence frequencies, sentence length in tokens (for LLMs) or words (for humans and baseline models). Random effects included the item number intercept and item number by plausibility slope. For Datasets 2 and 3, the formula was simplified to account for dataset structure (i.e., no item type or voice predictors).

Continuous variables were normalized before fitting. We used dummy coding for plausibility, with "plausible" as the reference level, dummy coding for item type, with "AA" as the reference level, and sum coding for voice. The analysis was conducted using the *lme4* R package (Bates et al., 2014).

**Probing analyses.** To compare the performance of probing classifiers across LLM layers, we divided LLM layers into three same-sized groups: early, middle, and late. Within each layer group, we compared average probe performance to the ceiling value (probe trained on human ratings; see **Section 2.4**), as well as the linear trend within each layer group (i.e., whether classifier performance increases, decreases, or stays constant within that layer group).

In all analyses, the results were FDR-corrected for the number of models within each category (humans, LLMs, and baselines). For probing analyses, the results were additionally corrected for the number of classifiers used within each analysis (e.g., 5 for generalization across trial types; 5 classifiers x 5 LLMs = 25 comparisons). Analysis code and data files can be found on GitHub: https://github.com/carina-kauf/lm-event-knowledge.

# 3. Results

We report a variety of tests to establish whether pre-trained LLMs are sensitive to event plausibility. In our main test (**Sections 3.1** and **3.2**), we investigate whether LLMs systematically assign higher scores to the plausible sentence compared to the implausible sentence within the minimal pair. We compare LLM performance with human performance (whether crowdsourced plausibility scores are higher for plausible than for implausible sentences within each pair) and with baseline model performance. Then we move beyond the minimal pair setup to conduct detailed analyses of all sentence scores, in order to determine the relative contributions of event plausibility and surface-level properties to LLM sentence scores (**Section 3.3)**. We investigate whether the event knowledge acquired by LLMs is *generalized* and *systematic* (**Section 3.4**), conduct an error analysis of LLM performance (**Section 3.5**), and use a probing analysis to track the emergence of explicit event plausibility signatures across LLM layers (**Section 3.6**).

## 3.1. All models show a gap between impossible and unlikely events

Our main sentence set (Dataset 1) contains two types of plausible-implausible sentence pairs: AI (animate-inanimate interactions, e.g., *The teacher bought the laptop* vs. *The laptop bought the teacher*) and AA (animate-animate interactions, e.g., *The nanny tutored the boy* vs. *The boy tutored the nanny*). In most cases, AI plausibility violations result in impossible events, whereas AA plausibility violations make the event unlikely but not impossible. We found that all language models exhibited differential performance on these sentence sets, with substantially better results for AI than for AA sentence pairs (**Figure 1**).

In the main analysis, we tested whether models systematically assign higher likelihood scores to plausible vs. implausible sentences within each minimal sentence pair. For each sentence pair, a model received a score of 1 if it assigned a higher score to the plausible version of the sentence and 0 otherwise. The same procedure was performed on human plausibility ratings for each sentence pair.

**AI sentence performance is high.** All models showed good performance on AI sentences (**Figure 1A, left**). MPT and RoBERTa scores were not significantly different from the human accuracy of 1, and other LLMs also had high performance, although slightly lower than humans (MPT: accuracy 0.97, $\chi 2=2.29$, $p=0.145$; GPT-J: accuracy 0.93, $\chi 2=7.37$, $p=0.011$; GPT-2: accuracy 0.95, $\chi 2=4.27$, $p=0.049$; RoBERTa: accuracy 0.98, $\chi 2=1.35$, $p=0.245$; BERT: accuracy 0.95, $\chi 2=4.27$, $p=0.044$). Baseline model performance was above chance, although not as high as that of LLMs and significantly lower than human performance; the best-performing baseline model was SDM, which was designed specifically to capture thematic fit for agent-verb-patient triplets (tinyLSTM: accuracy 0.80, $\chi 2=25.53$, $p<.001$; SDM: accuracy 0.90, $\chi 2=11.66$, $p<.001$; thematicFit: accuracy 0.73, $\chi 2=36.93$, $p<.001$; syntax-PPMI: accuracy 0.66, $\chi 2=50.74$, $p<.001$).

**AA sentence performance is moderate.** On AA sentences, all LLMs still performed above chance (**Figure 1A, right)** but their performance was significantly below the human accuracy of 0.95 (MPT: 0.84, $\chi 2=13.57$; GPT-J: 0.75, $\chi 2=27.12$; GPT-2: 0.74, $\chi 2=29.73$; RoBERTa: 0.78, $\chi 2=22.04$; BERT: 0.77, $\chi 2=24.56$; all $p<.001$). All baseline models performed at chance except for thematicFit (accuracy 0.62), indicating that information about AA event plausibility is more difficult to extract from subject-verb-object co-occurrence patterns in natural language than information about AI event plausibility.

**The gap between AI and AA sentences is significant.** As shown in **Table 3**, humans, LLMs, and two of the baseline models all show a performance gap between AI and AA sentence sets. However, the size of the gap for the models (average 0.18 for LLMs, 0.23 for baseline models) is much larger than the one in humans (0.05), a result we discuss further in **Section 4.5**. LLMs and most baseline models show comparable performance on the passive voice versions of AI and AA sentences (**Figure S4**).

For completeness, we also test the models on a set of AA-control items from Dataset 1, for which both sentences in a pair describe a plausible event (e.g., *The cheerleader kissed the quarterback vs. The quarterback kissed the cheerleader*). As expected, in that case the models produced comparable scores for the two events within each pair, as did humans (**Figures S5, S6**).

**Model-human score correlations also reflect the AI-AA gap**. We directly correlate model scores with human ratings (**Figure S7**) and show that the correlation is moderate for AI sentences (mean LLM r=.59) and poor for AA sentences (mean LLM r=.17). Note, however, that we would not necessarily expect LLM scores to fully align with human plausibility judgments, given that the models' task is word-in-context prediction, not plausibility evaluation per se. Nevertheless, this analysis helps reveal dissociable contributions of plausibility and language-specific features on LLM sentence likelihood scores, which we explore further in **Section 3.3**.

**Scaling helps to partially bridge the AI-AA gap.** To investigate the effect of LLM size on performance in more detail, we tested an extended set of 7 unidirectional models (with MPT being the largest) on AI and AA minimal pair performance (**Table S5; Figure S8**). We found consistently high performance on AI sentences across all tested models (even the smallest, DistilGPT-2 and the base GPT-2). AA sentence performance increased steadily with model size, although, as noted above, the gap was not fully bridged even for the 30-billion-parameter MPT model.

**Quantitative analysis confirms the validity of the binary labels.** To ensure that our binary accuracy results reflect meaningful plausibility differences in human ratings, we compute the average difference between plausible and implausible sentence scores within each pair. This value can range from -1 to 1 (with 1 reflecting a situation where people rated all plausible sentences as completely plausible and all implausible sentences as completely implausible). The mean difference was 0.78 (SD=.18) for AI sentences and 0.38 (SD=.24) or AA sentences, confirming the validity of our binary labels (see **Section 3.5** for more details).

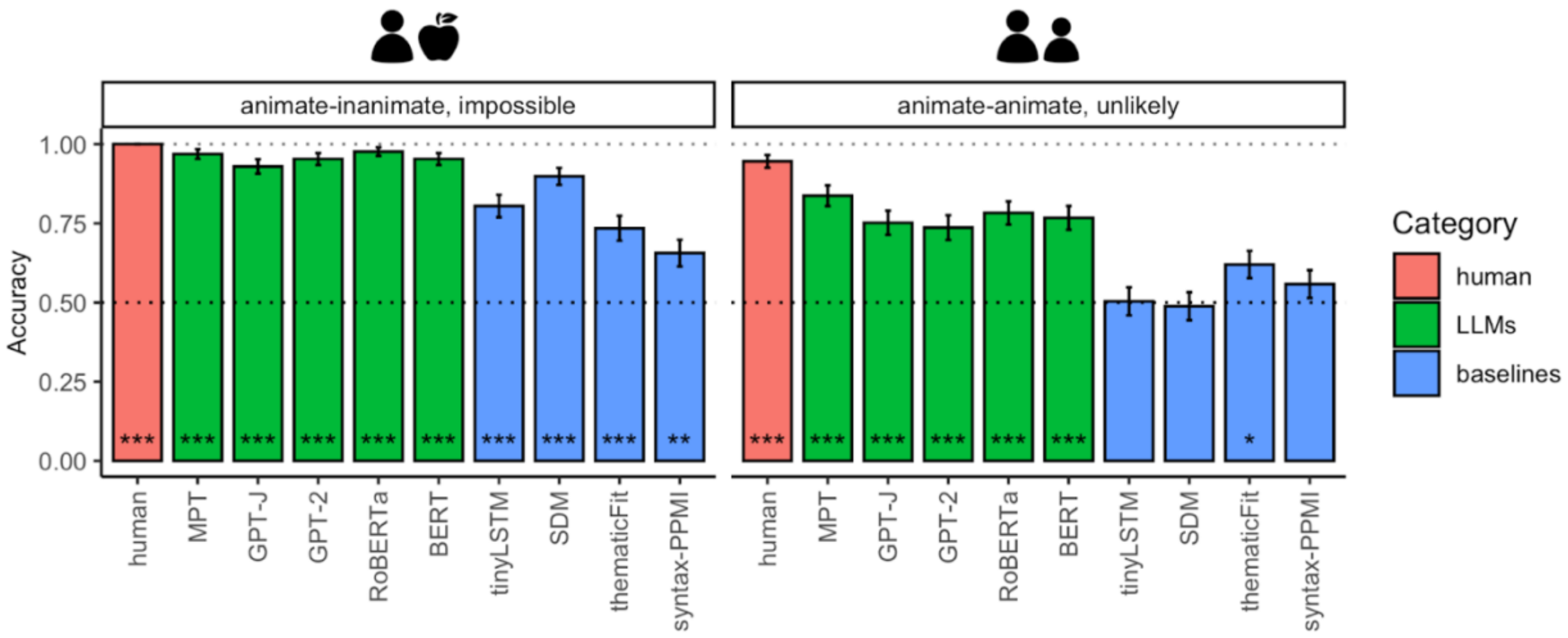

***Figure 1**. Main results, Dataset 1 sentences. Human, LLM, and baseline model accuracy scores for AI (left) and AA (right) sentence pairs. Significance was established via a binomial test. Here and elsewhere, significant results are marked with asterisks (p<0.05: *; p<0.01: **; p<0.001: ***). Error bars show the standard error of accuracy scores across sentence pairs.*

***Table 3.** Difference in performance between AI and AA sentence pairs.*

| Category | Metric | Difference | χ2 | p-value |
|---|---|---|---|---|
| human | human | 0.05 | 5.24 | 0.022 * |
| LLMs | MPT | 0.13 | 11.21 | <0.001 *** |
| | GPT-J | 0.18 | 13.84 | <0.001 *** |
| | GPT-2 | 0.22 | 21.34 | <0.001 *** |
| | RoBERTa | 0.19 | 20.92 | <0.001 *** |
| | BERT | 0.19 | 16.88 | <0.001 *** |
| baselines | tinyLSTM | 0.3 | 24.37 | <0.001 *** |
| | SDM | 0.41 | 48.84 | <0.001 *** |
| | thematicFit | 0.11 | 3.33 | 0.091 |
| | syntax-PPMI | 0.1 | 2.2 | 0.138 |

## 3.2. The gap in model performance between implausible and impossible events is not fully explainable by animacy or lexical variables

The gap between model performance on AI and AA sentences from Dataset 1 could be explained by several factors. First, implausible AI sentences in Dataset 1 mostly described impossible events (*The laptop bought the teacher*), whereas implausible AA sentences were often unlikely rather than impossible (*The boy tutored the nanny*), which resulted in a wider distribution of plausibility scores in humans (**Figure 3B**). Second, as follows from their name, AI sentences described animate-inanimate interactions, such that switching the agent and the patient typically violated the animacy selectional restriction on the verb; in contrast, AA sentences described animate-animate interactions, so our plausibility manipulation did not violate the animacy restriction. Finally, the AA sentences were more difficult overall (human accuracy 0.95 vs. 1 for AI sentences), possibly because AA sentences had a lower average word frequency (Google Ngram log frequency of 10.8 for AA vs. 11.1 for AI). To determine whether the latter two factors might explain differential model performance, we compared model and human performance on two additional sentence sets.

### 3.2.1. Dataset 2 (based on Vassallo et al., 2018)

This sentence set describes animate-inanimate (AI) interactions; plausibility is manipulated by varying the patient (e.g., *The actor won the award* vs. *The actor won the battle*; **Table 2**). Unlike AI sentences in Dataset 1, implausible sentences here are simply unlikely rather than impossible.

This difference is reflected in the distribution of human judgments for this sentence set, which are less polarized than for AI sentences from Dataset 1 (mean difference 0.55; see **Figure S9** for details). If argument animacy determines model performance, their accuracy on Dataset 2 should be similarly high to that for AI sentences from Dataset 1. If, on the other hand, unlikely events are more challenging for the models to evaluate compared to impossible events, then models should perform better on AI sentences from Dataset 1.

All models scored above chance but significantly below human performance of 0.99 (MPT: 0.93, $\chi 2$=20.8; GPT-J: 0.89, $\chi 2$=40.5; GPT-2: 0.88, $\chi 2$=46.1; RoBERTa: 0.91, $\chi 2$=29.5; BERT: 0.86, $\chi 2$=55.3; all p<.001). Average LLM performance on this sentence set (0.89) is higher than on AA sentence pairs from Dataset 1 (0.78) but is lower than on possible-impossible AI sentence pairs from Dataset 1 (0.96) (**Figure 2**). We therefore conclude that distinguishing likely and unlikely events remains a non-trivial challenge for LLMs even for AI sentences.

Further, the words in Dataset 2 are on average more frequent (log word frequency for Dataset 2: 11.5; log word frequency for AI sentences in Dataset 1: 11.1). We conclude that word frequency cannot fully account for the performance gap either.

### 3.2.2. Dataset 3 (based on Ivanova et al., 2021)

Dataset 3 is a small sentence set from a neuroimaging study by Ivanova et al. (2021) with the same manipulation as in Dataset 1: implausible sentences are generated by switching the agent and the patient (*The cop arrested the criminal* vs. *The criminal arrested the cop*; **Table 2**). Both agents and patients are animate. Average word frequency is higher than in Dataset 1 sentences (Google Ngram log frequency of 11.9), and human ratings are more polarized than those of AA sentences from Dataset 1 (mean difference for Dataset 3 = 0.76). Human accuracy for distinguishing plausible and implausible sentences in this dataset was 1, meaning that the plausibility judgments for this dataset were easy and unambiguous.

All models performed above chance but below human performance, who had perfect accuracy on this task, although the difference was non-significant for MPT and BERT (MPT: 0.89, $\chi 2$=2.37, n.s.; GPT-J: 0.82, $\chi 2$=5.66, p=.023; GPT-2: 0.84, $\chi 2$=4.52, p=.038; RoBERTa: 0.79, $\chi 2$=6.85, p=.014; BERT: 0.89, $\chi 2$=2.37, n.s.). Similar to Dataset 2, average LLM performance on this sentence set (0.85) falls between performance on AI sentences from Dataset 1 (0.96) and on AA sentences from Dataset 1 (0.78) (**Figure 2**). Although this dataset is too small to draw definitive conclusions, the results suggest that the performance gap between impossible and unlikely events in Dataset 1 cannot simply be explained by the fact that that likely-unlikely sentence pairs were more challenging.

Together, the results from **Sections 3.2.1** and **3.2.2** suggest that although event participant animacy and word frequency contribute to model performance, they do not fully explain performance patterns. In particular, unlikely sentences (across animacy configurations) pose challenges for LLMs despite being easy for humans.

Baseline model performance on Datasets 2 and 3 follows similar patterns to LLMs (**Figure 2**). In the remainder of the paper, we focus on LLM performance; detailed analyses of baseline model performance can be found in **SI 3**.

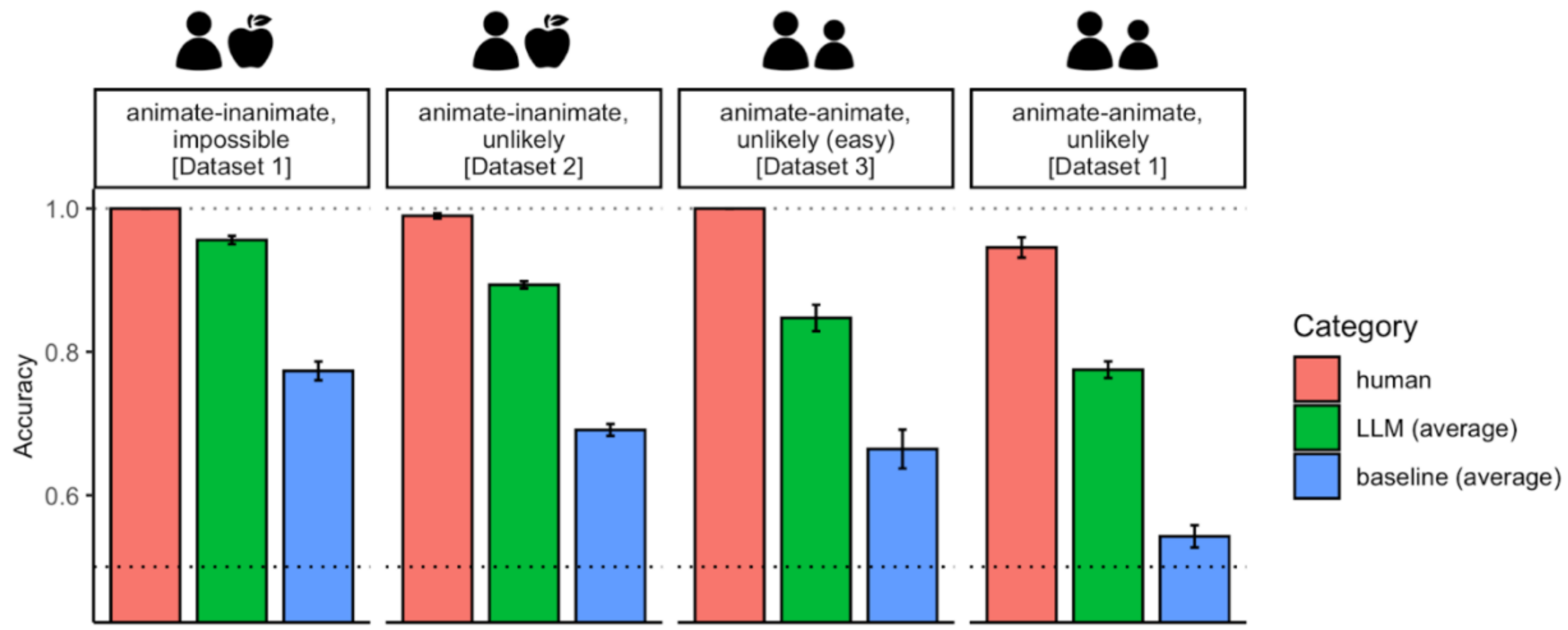

***Figure 2.*** *Human and model performance patterns on Dataset 1 (the first and last set of bars; same data as in **Figure 1**), as well as Datasets 2 and 3 (the second and third set of bars); results ordered by LLM performance. Dotted lines indicate chance-level performance.*

## 3.3. LLM scores are strongly influenced by surface-level sentence properties

So far, we have focused on comparing model scores within minimal pairs. Now we ask: to what extent do model scores dissociate for all plausible and implausible sentences in our datasets?

Under the view of LLMs as knowledge bases (Petroni et al., 2019), one might expect LLM scores to strongly track real-world plausibility, such that plausibility would be the main contributing factor to the probability of a sentence being generated. However, LLM outputs are known to be sensitive to diverse surface-level factors, most notably word frequency (e.g., Gong et al., 2018), which might overwhelm plausibility in determining the overall LLM score. Thus, we conduct a series of analyses to examine the relative contributions of plausibility and surface-level factors to the overall LLM score. As a control, we use human plausibility scores, which we expect to be primarily determined by plausibility and not surface-level properties of the stimulus.

**3.3.1. Plausible and implausible score distributions in language models show substantial overlap**. As shown in **Figure 3A**, human plausibility rating distributions for plausible and implausible sentences in Dataset 1 show little overlap (mean difference for AI sentences = 0.78, AA sentences = 0.38). In contrast, the distributions of likelihood scores assigned to plausible versus implausible sentences under language models show significant overlap (mean difference for LLMs: AI = 0.19, AA: 0.06; for baseline models: AI = 0.09, AA: 0.01). This suggests that language model scores are determined predominantly by factors other than plausibility, such as word frequency and sentence length.

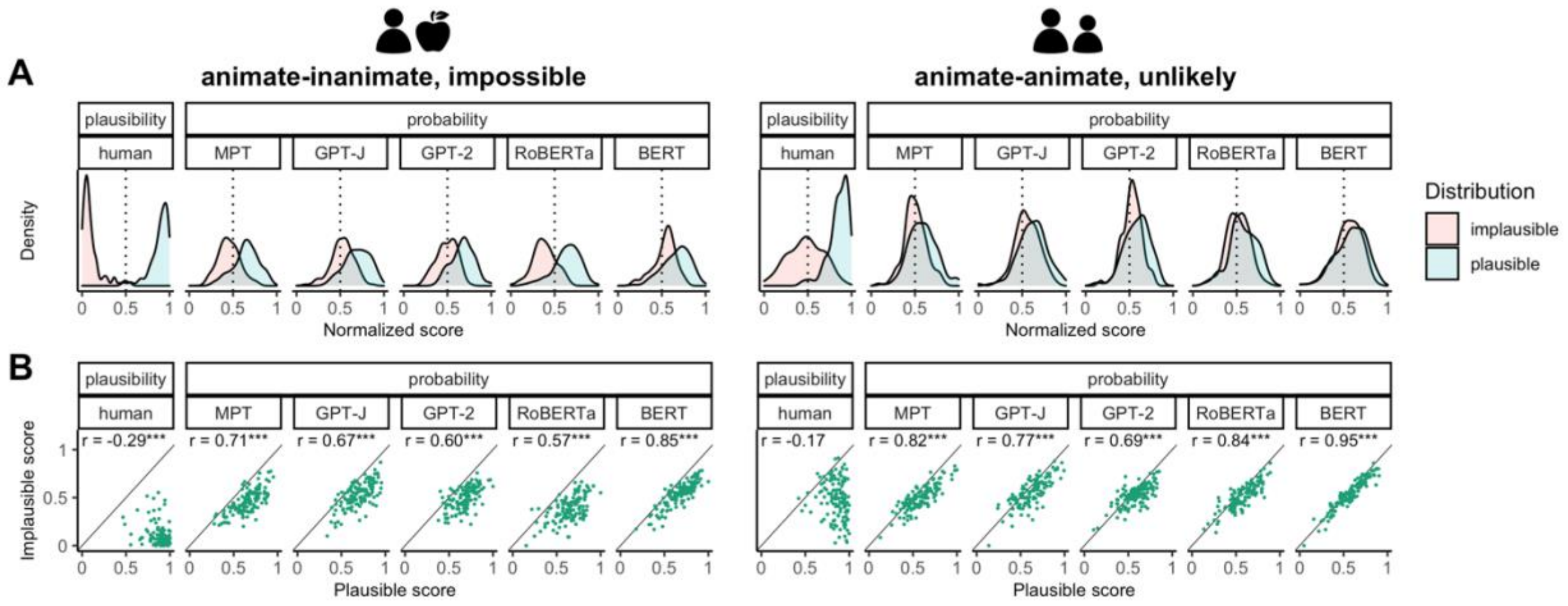

***Figure 3.*** *Human plausibility rating and model probability score distributions.* ***(A)*** *Density plots for plausible and implausible sentences. The dotted line shows the midpoint on the normalized score scale (0.5).* ***(B)*** *Correlation plots for plausible and implausible sentences. Each dot represents a sentence score. The diagonal is an identity line. Annotations show Pearson r correlation values and significance levels. See* ***Figure S1*** *for detailed analyses of the score distributions for the baseline models.*

**3.3.2. Switching the agent and the patient strongly influences human plausibility judgments but not LLM scores**. Our plausibility manipulation (switching the agent and patient in a sentence) was specifically designed to alter the plausibility of the described event while preserving the identities of individual words. If LLM scores depend on word-level properties (such as word frequency), the correlation between the scores for the two versions should be positive.

As shown in **Figure 3B**, human plausibility judgments show a negative correlation for plausible and implausible versions of the same AI sentence ($r=-0.29$, $p<.001$) and a non-significant correlation for AA sentences ($r=-0.17$, $p=.06$), indicating that word-level properties do not influence the average plausibility rating of a sentence pair. Conversely, under all LLMs, sentence likelihood scores exhibit a strong positive correlation (ranging from 0.57 for RoBERTa on AI sentences to 0.95 for BERT on AA sentences), indicating that LLM scores are largely driven by individual word features, rather than by event plausibility. This trend is more pronounced for AA than AI sentences, presumably due to a smaller relative contribution of plausibility – a hypothesis we explore next.

**3.3.3. Both plausibility and surface-level features predict LLM scores: mixed effects modeling.** To systematically test how different factors contribute to individual sentence scores, we fitted mixed effects models to likelihood scores from each model and to human plausibility judgements (**Table 4**; see Methods for model and contrast definition). Because we normalize the scores for each metric (humans and models), the resulting coefficients can be interpreted as effect sizes.

As expected, human plausibility ratings are primarily driven by the plausibility manipulations. For plausibility, we consider two main contrasts: (a) implausible AI sentences (*The laptop bought the teacher*) vs. implausible AA sentences (*The boy tutored the nanny*) and (b) implausible AA sentences vs. plausible AA sentences (*The nanny tutored the boy.*). In humans, the effect of AI vs. AA plausibility violation (-.37) is as strong as the implausibility effect for AA sentences (-.38).

All LLMs are also sensitive to both plausibility effects when assigning string likelihood scores; however, these effects are much weaker than the effects in human plausibility judgments, and the implausible AI>implausible AA effect (-.13) is larger than the implausible AA>plausible AA effect (-.07), consistent with the performance gap that we observed for AI and AA sentences.

In addition, model probabilities but not human plausibility judgments are sensitive to the main effects of surface-level sentence properties. Each LLM's performance on the critical task is affected by at least three of the following factors: voice, agent frequency, patient frequency, average word frequency, and sentence length, whereas human plausibility judgments are not affected by any of these features.

***Table 4****. Mixed effects modeling results. Variables that have a significant effect on human plausibility judgments are highlighted in bold. Sentence length is measured in tokens for LLMs, in words for humans. See* ***Table S2*** *for baseline model results. See* ***Tables S6*** *and* ***S7*** *for the same analysis for Datasets 2 and 3.*

| | | plausibility | probability | | | | | |
|---|---|---|---|---|---|---|---|---|
| | | human | MPT | GPT-J | GPT-2 | RoBERTa | BERT | Mean across LLMs |
| **Core effects** | **Implausible AA > Plausible AA** | **-0.38 *** ** | **-0.08 *** ** | **-0.07 *** ** | **-0.06 *** ** | **-0.07 *** ** | **-0.04 *** ** | **-0.07** |
| | **Implausible AI > Implausible AA** | **-0.37 *** ** | **-0.11 *** ** | **-0.11 *** ** | **-0.11 *** ** | **-0.2 *** ** | **-0.12 *** ** | **-0.13** |
| Surface -level effects | Voice | | | | | -0.06 *** | -0.13 *** | -0.04 |
| | Agent frequency | | 0.03 *** | 0.03 *** | 0.02 *** | | -0.01 * | 0.01 |
| | Patient frequency | | 0.03 *** | 0.03 *** | 0.02 *** | | -0.01 * | 0.02 |
| | Verb frequency | | | | | | | 0 |
| | Avg. word frequency | | | | | 0.03 ** | | 0.01 |
| | Sentence length | | -0.02 *** | -0.02 *** | -0.02 *** | -0.03 *** | -0.07 *** | -0.03 |
| | Voice x Sentence (AA>control) | | | | | | | 0.01 |
| | **Voice x Sentence (AI>AA)** | **0.03 ** ** | | | **0.03 ** ** | **0.04 *** ** | | **0.02** |
| | Plausibility x Voice x Sentence (AA>control) | | | | | | | 0 |
| | **Plausibility x Voice x Sentence (AI>AA)** | **-0.07 *** ** | | | | | **0.04 *** ** | **0.01** |

Finally, even in humans, the AI implausibility effect is modulated by some surface-level properties. Compared to AA sentences, humans are likely to assign more polarized scores to AI sentences presented in active voice than in passive voice (higher for plausible, lower for implausible). GPT-2 and RoBERTa likelihood scores partially capture this effect, and BERT shows an effect in the opposite direction, penalizing passive implausible AI sentences more harshly. The best performing LLM, MPT, fails to capture the fine-grained effects of surface-level properties on human judgments.

Overall, the mixed-effects model analysis is consistent with other analyses. All LLMs show a significant effect of plausibility on resulting sentence likelihood scores, indicating that they are sensitive to generalized event knowledge. Yet, we still observe a performance gap between AA and AI sentences and a strong effect of surface-level linguistic properties on LLM sentence scores that diverge from those of human plausibility judgments, indicating that raw probability of an event description cannot be used directly as an indicator of its plausibility.

## 3.4 LLMs generalize well across syntactic sentence variants, but only partially across semantic sentence variants

In the previous section, we demonstrated that LLM sentence scores are strongly influenced by surface-level sentence features, a factor that might negatively affect these models' ability to generalize. In this section, we directly tested how well LLM scores generalize across different forms of a sentence. To do so, we evaluated the extent to which model scores generalize across sentence voice (active vs. passive) and across sentences with synonymous, or closely related, meanings.

**3.4.1. LLMs generalize across active and passive sentences**. To test invariance to sentence syntax, we calculated the Pearson correlations between the active and passive voice versions of the same sentence (*The teacher bought the laptop* vs. *The laptop was bought by the teacher*; **Figure 4A**). Human scores were highly correlated (r=0.96, p<.001), indicating that human plausibility ratings are indeed invariant to sentence voice. LLM likelihood scores were also strongly correlated (max: BERT, r=.93; min: GPT-J/GPT-2, r=.79; all p<.001), indicating that LLMs can successfully generalize across active and passive voice forms of the same sentence.

**3.4.2. LLMs show some generalization across synonymous sentences**. To test invariance to specific lexical forms, we compared scores for sentence pairs where subject, verb, and object words were synonymous, or closely related in meaning (*The teacher bought the laptop* vs. *The instructor purchased the computer*; **Figure 4B**). Human judgments were highly correlated across synonymous sentence pairs (r=.90, p<.001), indicating that they are largely invariant to specific word identity. LLMs showed some generalization (max: MPT and RoBERTa, r=.56; min: BERT, r=.27; all p<.001), indicating that these models are somewhat consistent in assigning scores to synonymous utterances, but this relationship is far weaker than that observed in humans or than LLMs' syntactic generalization capabilities. This result is consistent with the results in **Section 3.3.3**, which showed that the models are sensitive to lexical-item-level properties, such as word frequency, and presents a potential challenge for robust representations of generalized event knowledge in LLMs.

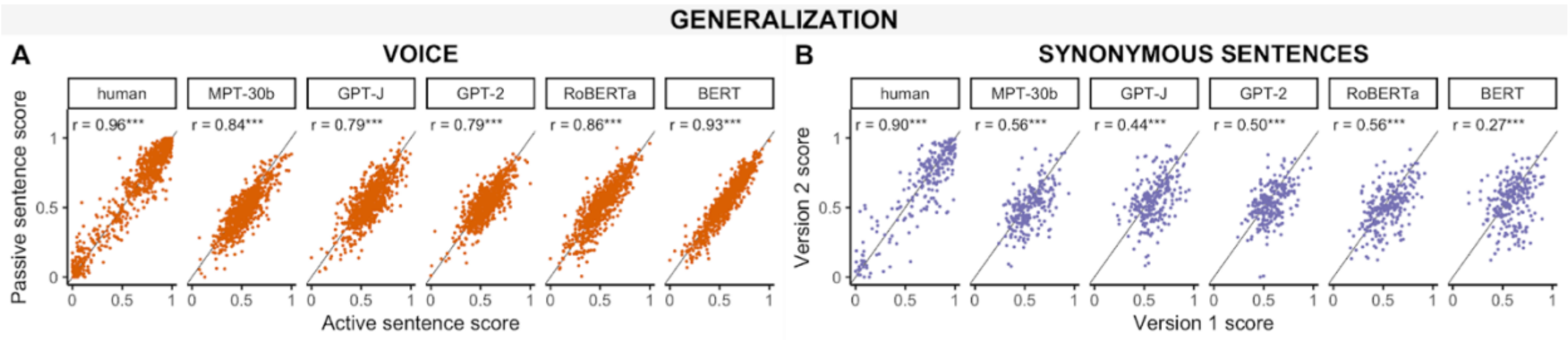

***Figure 4**. Generalization results. **(A)** Human and LLM scores for active voice and passive voice versions of the same sentence (e.g., 'The author finished the novel.' vs. 'The novel was finished by the author.'). **(B)** Human and LLM scores for sentences that have synonymous, or closely related, meanings (e.g., 'The author finished the novel.' vs. 'The writer completed the book.'). Each dot represents a sentence score. The diagonal is an identity line. See **Figure S2** for baseline model results.*

# 3.5. LLM deviations from ground-truth labels are partially, but not fully explained by plausibility violation strength

To understand the nature and severity of LLM errors, we conducted a quantitative and a qualitative analysis of the sentence pairs that most LLMs got wrong.

We first tested whether the severity of the plausibility violation correlates with model performance. To do so, we correlated the violation magnitude in each sentence pair (operationalized as the difference between human scores for plausible and implausible sentence versions) and the number of LLMs (0 through 5) that correctly evaluated that sentence pair. For both AI and AA sentences, we observed a moderate positive correlation, suggesting that sentence pairs that are more difficult for humans to decide are also more challenging for LLMs.

Then, we conducted a qualitative analysis of sentence pairs that most LLMs got wrong (**Table 5**). We found that these include several sentence pairs where human judgments actually deviated from ground truth labels (e.g., *The orderly assisted the dentist* vs. *The dentist assisted the orderly*; see **Table S4**), but in two-thirds of the cases there was at least a 0.1 difference between plausible and implausible sentence ratings in humans. Some errors might be explained by low-level features of the input such as non-standard spelling (e.g., *tour-guide* instead of *tour guide*); some might be caused by low-frequency words (e.g., *milliner*) that were underrepresented in the models' training data; and some might reflect a failure to identify typical agent/patient roles (e.g., most LLMs fail to identify *trainee* as a typical patient for the verb *taught*, even though human judgments in this example are rather unambiguous). Overall, we conclude that (1) many of the models' errors are "reasonable", being caused by ambiguous event plausibility labels and non-standard spelling; (2) the knowledge gap for unlikely (AA) sentences cannot be fully explained by such "reasonable" errors.

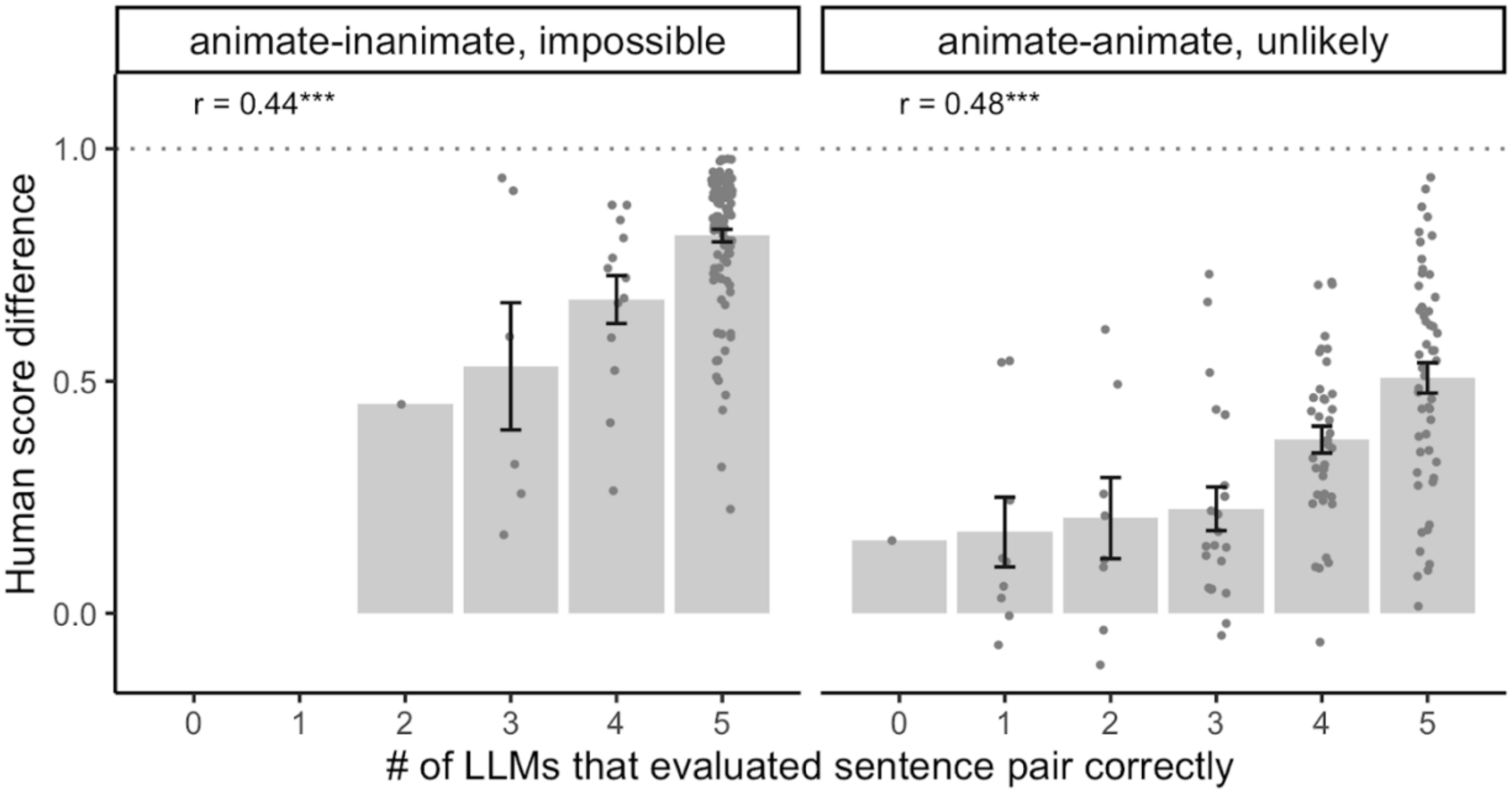

***Figure 5***. *Error analysis for Dataset 1. The number of LLMs (out of 5) which evaluated a given sentence pair correctly correlates with the magnitude of the human score difference between plausible and implausible versions of that sentence (the higher the difference, the better humans are at distinguishing plausible and implausible sentence versions). Each dot is a minimal sentence pair; error bars denote standard errors of the mean. See* ***Figure S3*** *for baseline model results. See* ***Figure S10*** *for the same analysis for Datasets 2 and 3.*

***Table 5***. *All sentence pairs (out of 391) that were evaluated correctly by at most 1 LLM, ordered by human score difference from largest to smallest. Sentences where the human ratings also deviated from the ground truth labels are grayed out. See* ***Table S3*** *for baseline model results. See* ***Tables S8*** *and* ***S9*** *for the same analysis for Datasets 2 and 3.*

| | Trial type | #LLMs correct (of 5) | Human score difference | Plausible sentence | Implausible sentence |
|---|---|---|---|---|---|
| 1 | AA | 2 | 0.61 | The principal scolded the child. | The child scolded the principal. |
| 2 | AA | 1 | 0.54 | The craftsman taught the trainee. | The trainee taught the craftsman. |
| 3 | AA | 1 | 0.54 | The lion chased the tour-guide. | The tour-guide chased the lion. |
| 4 | AA | 2 | 0.49 | The brunette tipped the busboy. | The busboy tipped the brunette. |
| 5 | AI | 2 | 0.45 | The milliner adorned the fedora. | The fedora adorned the milliner. |
| 6 | AA | 2 | 0.26 | The impersonator conned the inspector. | The inspector conned the impersonator. |
| 7 | AA | 1 | 0.24 | The vagabond revered the priest. | The priest revered the vagabond. |
| 8 | AA | 2 | 0.21 | The deceiver imitated the conqueror. | The conqueror imitated the deceiver. |
| 9 | AA | 0 | 0.16 | The environmentalist cautioned the tobacconist. | The tobacconist cautioned the environmentalist. |
| 10 | AA | 1 | 0.12 | The warmonger terrorized the gunsmith. | The gunsmith terrorized the warmonger. |
| 11 | AA | 2 | 0.12 | The biker defied the trainer. | The trainer defied the biker. |
| 12 | AA | 1 | 0.11 | The nomad cherished the clergyman. | The clergyman cherished the nomad. |
| 13 | AA | 2 | 0.1 | The genius shocked the cousin. | The cousin shocked the genius. |
| 14 | AA | 1 | 0.06 | The prodigy surprised the relative. | The relative surprised the prodigy. |
| 15 | AA | 1 | 0.03 | The neuroscientist overwhelmed the lab assistant. | The lab assistant overwhelmed the neuroscientist. |
| *16* | *AA* | *1* | *-0.01* | *The liar emulated the victor.* | *The victor emulated the liar.* |
| *17* | *AA* | *2* | *-0.04* | *The reviewer criticized the right-winger.* | *The right-winger criticized the reviewer.* |
| *18* | *AA* | *1* | *-0.07* | *The pixie mesmerized the ogre.* | *The ogre mesmerized the pixie.* |
| *19* | *AA* | *2* | *-0.11* | *The orderly assisted the dentist.* | *The dentist assisted the orderly.* |

## 3.6 Internal representations of event plausibility generalize across sentences

The previous sections have investigated the behavioral performance of LLMs in distinguishing plausible and implausible events. Here, we ask: is the distinction between plausible and implausible events encoded in the LLMs' representational spaces? Can a linear classifier trained to distinguish plausible and implausible events generalize to new sentences? If so, an LLM may have learned to represent plausible and implausible events in systematically different ways, a strategy that might help it to generalize in spite of its sensitivity to surface-level properties.

We find that sentence plausibility is indeed linearly decodable from internal LLM representations (**Figure 5**; **Table 6**). Consistent with our main results (**Section 3.1**), impossible AI sentences have a much stronger plausibility signature than unlikely AA sentences, with AI-to-AI classifier performance reaching ceiling for late and, in some cases, middle model layers, whereas AA-to-AA classifier performance  on most LLM representations reaches above-chances levels later and (except for certain middle layers in MPT) generally falls short of the ceiling (**Figure 6A**). The successful performance of both classifiers indicates that plausibility is one of the organizing dimensions of the underlying distributional spaces for middle and late layers.

The fact that sentence representations of later model layers are more suitable for decoding plausibility than those of earlier layers is consistent with previous results showing that semantic information tends to be encoded more strongly in later layers (Belinkov et al., 2017; Tenney et al., 2019; Papadimitriou et al., 2022). The trend we observed in one of the models, MPT, where mid-layer performance exceeded late-layer performance, should be examined further in other large (multi-billion parameter) models.

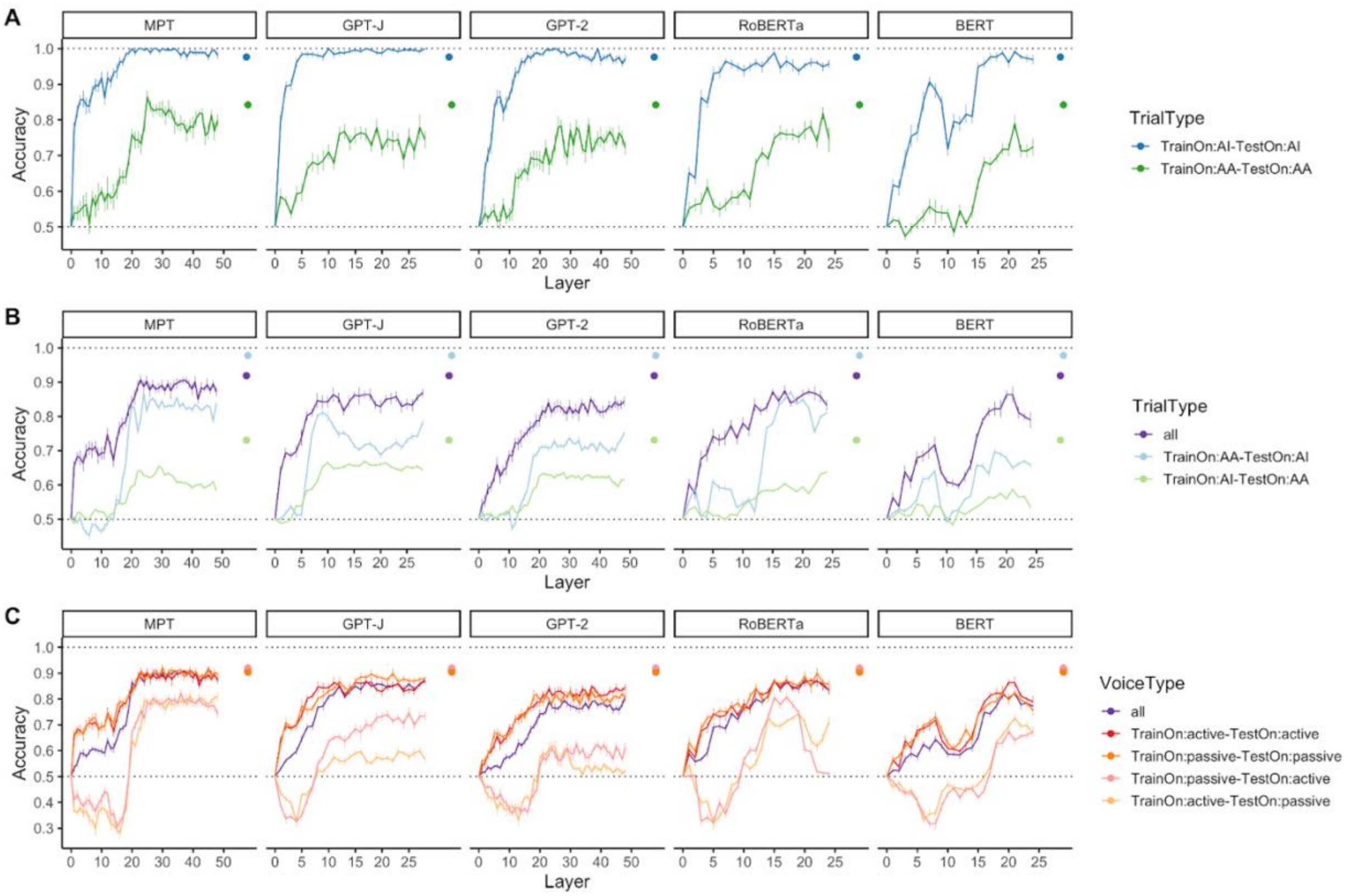

***Figure 6.*** *Classification accuracies for linear probes trained to differentiate plausible from implausible event descriptions in model embeddings. (**A**) Performance within an item type (animate-inanimate, AI, vs. animate-animate, AA), active voice. (**B**) Generalization across item type, active voice. (**C**) Generalization across active and passive voice. The dots denote classification accuracy of probes trained on human scores (which can serve as an empirical ceiling value). Dotted lines indicate chance-level and ideal performance. Error bars show the standard error of the mean across the 10 cross-validation folds.*

Next, we examined how well the plausibility signature generalizes across different sentence types (AI vs. AA) and across different surface-level forms (active vs. passive).

Generalizing across sentence types is only partially successful. Generalizing to AA sentences from AI sentences leads to a drop in classifier accuracy compared to testing an AI-trained classifier on AI sentences (**Figure 6B**). In contrast, probes that are trained to distinguish plausible vs. implausible AA sentences have similar performance on AI and AA test sets, although they fall short of the probes trained and evaluated on sentence representations from both sentence sets (labeled “all” in the figure).

Generalizing across sentence voice is possible given the right training regime. A classifier trained and evaluated on all sentences (active and passive) performs as well as classifiers trained and evaluated on sentences in only one voice (**Figure 6C**), a result consistent with good voice generalization performance in **Section 4.4**. However, active-to-passive and passive-to-active classifiers perform substantially worse, with below-chance accuracy in early layers and sometimes at-chance accuracy in late layers, indicating that these classifiers leverage surface-level plausibility signatures that do not generalize across surface forms. Thus, LLM sentence embeddings contain both syntax-specific and syntax-invariant plausibility information.

Overall, from the probing results we conclude that event plausibility is linearly decodable from LLM sentence embeddings. This plausibility information becomes salient in middle LLM layers and remains high thereafter, making it possible to use this information during output generation. Similar to the behavioral results, there is a performance gap between AI and AA sentences, and the plausibility signature only partially generalizes across AI and AA event types (although it can generalize across sentence voice).

For detailed statistical comparisons, see **Tables S10** and **S11**. For extended probing results, see **Figures S13** and **S14**; **Table S12**.

# 4. Discussion

Can generalized event knowledge emerge from distributional information encoded in language? To find out, we compared the likelihood scores that pre-trained LLMs assigned to plausible vs. implausible event descriptions. To minimize the putative influence of confounding factors, we used syntactically simple, tightly controlled minimal pair sentences. We demonstrate that LLMs acquire substantial event knowledge and improve over strong baseline distributional models, especially when it comes to distinguishing possible and impossible events (*The teacher bought the laptop* vs. *The laptop bought the teacher*). However, they are less consistent when assigning probabilities to likely events vs. events that are unlikely but not impossible (*The nanny tutored the boy* vs. *The boy tutored the nanny*). Using three different sentence sets, we demonstrated that this gap in performance cannot be fully explained by the animacy of the event participants or word frequency.

We further conducted a rigorous set of analyses to elucidate the relationship between an LLM sentence score (which reflects its generation probability) and plausibility, showing that LLM scores depend both on sentence plausibility and surface-level sentence properties. In generalization analyses, we found that both LLM and human scores are consistent for active and passive voice versions of the same sentence, but LLMs are less consistent than humans for synonymous sentence forms. Lastly, we found that sentence plausibility is linearly decodable from internal LLM representations, with the same gap between impossible and unlikely event performance as that observed in behavioral tests. We conclude that sentence plausibility is a major contributor to sentence generation probability, although this relationship is less clear-cut for likely vs. unlikely event descriptions.

## 4.1 In case of impossible events, LLMs might leverage selectional restrictions

LLMs in our study were close to ceiling when distinguishing possible and impossible events. A notable feature of the impossible event descriptions in our datasets is the violation of selectional restrictions on the verb, i.e., the set of semantic features that a verb requires of its arguments (e.g., requiring an agent to be animate) (Katz & Fodor, 1963; Chomsky, 1965; Levin, 1993). When plausibility violations were not driven by selectional restrictions (as in the "unlikely" sentence sets), model performance dropped.

Our findings suggest that selectional restrictions are a linguistic property that is learnable from corpus data (as also confirmed by the large number of experiments with computational methods for selectional restriction acquisition from texts; e.g., Erk, 2007; Thrush et al., 2020) and whose violations are meaningfully distinct from violations of graded world knowledge (Warren & McConnell, 2007; cf. Matsuki et al., 2011). Computational evidence suggests that BERT models are able to generalize their knowledge of selectional restrictions in novel word-learning paradigms (Thrush et al., 2020) and can partially rely on the semantics of the head predicate to predict upcoming event participants (Metheniti et al., 2020). The asymmetry in performance on possible/impossible vs. likely/unlikely events was independent from the specifics of LLM architecture and training and was additionally present, in an even more marked way, in our baseline models. Furthermore, a classifier probe trained on possible vs. impossible sentence embeddings performed almost perfectly on other sentences from the same category but failed to generalize to likely vs. unlikely events, indicating that selectional restrictions have a distinct representational signature. These results are consistent with psycholinguistic evidence from reading times and EEG indicating that violations of selectional restrictions and violations of world knowledge evoke distinct processing signatures (e.g., Sitnikova et al., 2008; Paczynski & Kuperberg, 2012; Warren et al., 2015; cf. Hagoort et al., 2004).

The fact that selectional restrictions are easier to learn from distributional linguistic data than graded event likelihood is an important distinction, as both of these factors affect plausibility judgments in humans (e.g., Hagoort et al., 2004; Warren et al., 2015). To verify and extend our findings, future work should test LLMs' knowledge of selectional restrictions on features other than animacy, such as the physical constraints that a predicate places on its patients (Wang et al., 2018), evaluate their performance on impossible events that do not violate selectional restrictions per se (e.g., *She gave birth to her mother, The man was killed twice*, or *After 10 coin tosses, she got 12 heads.*), and conduct more targeted tests of agent-verb and patient-verb plausibility (Metheniti et al., 2020).

## 4.2. LLMs can infer thematic roles

The stimuli in Datasets 1 and 3 are constructed such that the model has to leverage word order information to successfully determine event plausibility. LLMs successfully accomplish this task for most possible vs. impossible events and for a number of likely vs. unlikely events. Furthermore, they produce highly correlated scores for active and passive versions of the same sentence, suggesting that thematic role information generalizes beyond a specific word order.

The probing results produce additional insight into the emergence of thematic role information in the LLMs (**Figure 5b**). A probe that is trained on a mix of active and passive sentences performs as successfully as the probe trained and tested on only one voice type, suggesting that plausible and implausible sentence embeddings in late LLM layers are linearly separable by the same hyperplane across syntactic structures. This finding aligns with recent computational work showing that even though most sentences in the language input describe prototypical events (Mahowald, Diachek, et al., 2023), LLMs are able to correctly represent the argument structure of non-prototypical event descriptions in late layers (Papadimitriou et al., 2022).

Despite LLMs' general success in thematic role inference, some confusion about thematic role assignment might remain for unlikely events in our main dataset. These events describe animate-animate (AA) interactions. Animate direct objects are treated as a special case in many languages. In Spanish, for example, they are differentially marked with a preposition (e.g., Bossong, 1991; Aissen, 2003). Even though English (the language that we test here) does not overtly mark any direct objects, it could be that the correct thematic role assignment remains ambiguous for AA sentences in a way that it does not for AI sentences. In humans, this ambiguity can lead to a reinterpretation of the sentence as plausible even when the word order indicates an implausible interpretation (e.g., Gibson et al., 2013); a fine-tuned conversational LLM, ChatGPT-3.5, exhibited a similar bias (Cai et al., 2023), suggesting that in LLMs, like in humans, plausibility priors might overrule thematic role assignment.

## 4.3 The 'reporting bias' in language corpora makes it harder to distinguish likely and unlikely events

A core challenge for modeling plausibility based on linguistic input is the fact that the frequency with which events are described in the language is not a reliable predictor of the frequency with which events occur in the real world. Because much of our world knowledge is shared across individuals (e.g., McRae et al., 2005) and human communication is shaped by efficiency (Gibson et al., 2019) and cooperation (Grice, 1975), language is biased towards reporting extraordinary facts and events rather than the trivial (Gordon & Van Durme, 2013). Many commonsense facts about the world are thus presupposed rather than stated explicitly; in contrast, unusual events are discussed extensively. As a result, likely events are underrepresented in linguistic corpora, whereas unlikely events are overrepresented.

The 'reporting bias' of rare and newsworthy events in language corpora has traditionally provided difficulty for modeling semantic knowledge via text mining (e.g., Lucy & Gauthier, 2017). Recent studies probing world knowledge in LLMs show that although the generalization capabilities of these models are able to overcome the reporting bias to some extent (Shwartz & Choi, 2020; Weir et al., 2020), they still tend to reflect biases that exist in their training corpus (Zmigrod et al., 2019;

Shwartz & Choi, 2020; Vig et al., 2020). As a result, one explanation of the performance gap that we observe for likely vs. unlikely events in LLMs could be that unlikely events are overrepresented in the corpus, leading the models to predict them as frequently as likely events. In contrast, impossible events are nearly absent from the training data, and so the models correctly assign them low likelihood scores.

It is worth noting that the reporting bias present in pragmatically influenced natural language also affects concept learning in humans: blind people's beliefs about the canonical color of animals (Kim et al., 2019), for example, are consistent with the inadequate color information encoded in sighted people's linguistic productions (Ostarek et al., 2019). This, along with successful acquisition of many other visual concepts by the blind (e.g., Marmor, 1978; Landau & Gleitman, 1985; Wang et al., 2020), implies that learning from distributional linguistic information is a likely, though not the only, strategy that humans adduce to organize facets of world knowledge, especially those to which they do not have direct sensorimotor access.

A possible solution to overcoming the reporting bias would be to adjust the event distribution via injecting manually elicited knowledge about object and entity properties into models (Wang et al., 2018; although see Porada et al., 2021) or via data augmentation (e.g., Zmigrod et al., 2019). Alternatively, information about event typicality might enter LLMs through input from different modalities, such as visual depictions of the world in the form of large databases of images and/or image descriptions (Bisk et al., 2020). Distributional models trained on multimodal data have indeed been shown to outperform text-only trained models in overcoming the reporting bias for visual concept knowledge (e.g., Paik et al., 2021; Zhang et al., 2022). In the future, we plan to extend our analysis of GEK to multimodal LLMs (e.g., CLIP; Radford et al., 2021) in order to investigate the role of extralinguistic evidence, which might reduce the impact of the reporting bias and better simulate the multimodal information that humans use to acquire GEK. Finally, a training objective that emphasizes robust, generalizable event representations might lead to more robust GEK knowledge than word-in-context prediction, although what such an objective would look like remains to be discovered.

## 4.4 Distributional language models are good models of language but imperfect models of world knowledge

We have shown that the probability for generating a particular sentence under a given LLM depends both on plausibility and on surface-level features of that sentence, such as word frequency. This result is largely expected, because distributional models are naturally geared toward producing more frequent tokens more often. However, it does result in a high overlap between the score distributions we observe for plausible and implausible sentences, meaning that many implausible sentences have higher likelihood generation simply because they contain frequent words.

The fact that LLMs are sensitive to both sentence plausibility and surface-level features makes them good candidate models of human language processing. On the one hand, sentence plausibility substantially facilitates language processing in humans (e.g., Kutas & Hillyard, 1984; Federmeier & Kutas, 1999; McRae & Matsuki, 2009; Bicknell et al., 2010). On the other hand, humans are also sensitive to lexical frequency effects when processing linguistic inputs (e.g., Broadbent, 1967; Rayner & Duffy, 1986; Goodkind & Bicknell, 2021; Haeuser & Kray, 2022) and can use both linguistic knowledge and event knowledge in real time depending on task demands (Willits et al., 2015). As a result, LLM scores are a good predictor of human reading times (Oh et al., 2022; Shain et al., 2022; Oh & Schuler, 2023), neural predictability signatures like N400 (Michaelov et al., 2023; Szewczyk & Federmeier, 2022), and brain response patterns to individual sentences (e.g., Schrimpf et al., 2021; Caucheteux & King, 2022; Tuckute et al., 2023).

However, sensitivity to surface-level features of the input can make LLMs unreliable as knowledge bases. Due to this sensitivity, they produce inconsistent results when the same description is phrased differently (Ravichander et al., 2020; Ribeiro et al., 2020; Elazar et al., 2021a), produce unsystematic judgments (Talmor et al., 2020), hallucinate facts (Ji et al., 2023; Liu et al., 2022), fail to learn commonsense event schemas (Pedinotti et al., 2021), and generalize only weakly across synonymous descriptions of the same event (**Section 4.4**). The ability to abstract away from specific inputs is a key feature of GEK; thus, the ability of future language-based models to acquire robust, flexible event schemas will depend crucially on their ability to generalize beyond corpus statistics.

Even though world knowledge and language processing behavior are closely linked in humans, world knowledge and language are two fundamentally different capabilities that have been shown to dissociate in humans (e.g., Caramazza et al., 1982; Patterson et al., 2007; Lambon Ralph et al., 2017), including in a study that specifically evaluated event plausibility (Ivanova et al., 2021). We therefore speculate that acquisition of robust, statistics-invariant world knowledge representations would require a different objective function from that required for acquiring linguistic proficiency (Mahowald, Ivanova et al., 2023). The word-in-context prediction objective, which enables LLMs to excel at acquiring formal linguistic competence, encourages pre-trained LLMs to organize their semantic spaces mainly by relatively simple features such as similarity and association (Lenci, 2023). This organization principle, however, does not always lead to robust concepts and relations, which are useful for natural language understanding tasks and serve as important units for developing more complex semantic structures (Lenci, 2023; Lenci & Sahlgren, 2023).

Based on our results and on studies from the literature, we conclude that the word-in-context prediction objective alone is suitable for acquiring a wealth of event knowledge but cannot ensure the consistency of these representations. Thus, in both humans and models, distributional linguistic knowledge is not a replacement for GEK but rather a useful foundation for further enrichment and fine-tuning of generalized semantic representations.

## 4.5 Generating descriptions of unlikely events: a feature rather than a flaw?

The fact that LLMs' distributional linguistic knowledge does not limit them to the realm of plausible events could be considered a feature rather than a flaw. The power of language is not only in its ability to convey factual knowledge: language allows humans to brainstorm, fantasize, discuss counterfactuals, speculate, and dream. With enough backstory, even an impossible event like *The laptop bought the teacher* can be rendered plausible, eliminating the processing difficulty in humans (e.g., Nieuwland & Van Berkum, 2006; Warren et al., 2008; Jouravlev et al., 2019) and in LLMs (Michaelov et al., 2022). Thus, restricting the models to the realm of *a priori* plausible events would handicap their potential as models of human language. Of course, in the absence of contextual information (as is the case in our study), we would still expect LLMs to generate plausible event descriptions more often than implausible ones. However, an overly strong alignment between an LLM and a knowledge base will likely be counterproductive for its linguistic fluency.

Finally, a naïve approach to pre-trained LLMs as knowledge bases overlooks their core design feature: they are prediction machines that aim to faithfully mimic all properties of the input, not simply semantic plausibility. As seen in **Section 3.3**, LLM scores are sensitive to a variety of surface-level properties of the stimulus that need to be factored out to receive a more faithful estimate of plausibility. Therefore, LLMs should be regarded at most only as partial models of human semantic plausibility. In turn, if the goal is to directly compare LLM scores with human scores, one should consider a human metric that is more appropriate, such as reading times (Oh et al., 2022).

Prediction-based LLMs are an important tool for investigating which cognitive capacities can, in principle, rely on distributional linguistic knowledge. Contemporary LLMs show that large amounts of world knowledge can be learned from language alone with a simple word-in-context prediction objective, yet controlled, targeted manipulations like the ones used in this study can also highlight areas of knowledge where LLM behavior is not yet fully aligned with human behavior. Future work should explore the extent to which LLMs master other types of event knowledge, such as knowledge of typical/possible event sequences, knowledge of impossible events that do not violate selectional restrictions per se, and the extent of their sensitivity to selectional restrictions other than animacy. Furthermore, the fact that LLMs in our study sometimes perform below humans even on syntactically simple sentences (*The X Ved the Y*) suggests that testing them on longer sequences of text might uncover even larger deviations from GEK. Overall, detailed investigations of world knowledge in distributional language models are a valuable source of evidence for clarifying the relationship between language and broader cognition.

# Acknowledgments

We thank Josh Tenenbaum, Roger Levy, Jacob Andreas, and HuthLab members for helpful comments. This collaborative work was made possible thanks to the MIT-UNIPI Project, a grant from the MISTI Global Seed Fund (the MIT-Italy program). CK was supported by the K. Lisa Yang Integrative Computational Neuroscience (ICoN) Center at MIT. AI was supported by the Whitaker

Health Sciences Fund Fellowship from MIT and by MIT Quest for Intelligence. GR contributed to this work during her PhD granted by the University of Pisa. EC was supported by the General Research Fund "Modeling Generalized Event Knowledge for Noun Compound Interpretation and Prediction with Vector Spaces and Transformers" (B-Q0AH). EF was supported by NIH awards R01-DC016607, R01-DC016950, and U01-NS121471, as well as by research funds from the McGovern Institute for Brain Research, the Brain and Cognitive Sciences department, the Simons Center for the Social Brain, and the Middleton Professorship. This research was also partly funded by PNRR - M4C2 - Investimento 1.3, Partenariato Esteso PE00000013 – "FAIR - Future Artificial Intelligence Research" - Spoke 1 "Human-centered AI", funded by the European Commission under the NextGeneration EU programme.

# Supplemental Information

## SI1. LLM design features

***Table S1***. *Overview of LLM designs. BPE = BytePair Encoding, CLM= Causal Language Modeling, MLM = Masked Language Modeling. NSP = Next-Sentence Prediction*

| Model | Attention | Tokenization | #parameters | Vocabulary size | Training data size | Training task |
|---|---|---|---|---|---|---|
| **mpt-30b** | Unidirectional | BPE | 30B | 50K | 1TB | CLM |
| **gpt-J-6b** | Unidirectional | BPE | 6B | 50K | 800GB | CLM |
| **gpt2-xl** | Unidirectional | BPE | 1.5B | 50K | 40GB | CLM |
| **roberta-large** | Bidirectional | WordPiece | 355M | 30K | 160GB | Dynamic MLM |
| **bert-large-cased** | Bidirectional | WordPiece | 340M | 30K | 13GB | MLM + NSP |

## SI2. Baseline model description details

**Baseline models.** We are interested in investigating whether knowledge of event plausibility emerges as a natural by-product of attending to word co-occurrence statistics. As a result, we compare the performance of the LLMs against four baseline models designed to encode relevant information for building an accurate event representation from linguistic input.

The **TinyLSTM** model is a vanilla two-layer LSTM recurrent neural network, trained with a next-word prediction objective on the string data from the 1-million-word English Penn Treebank §2-21 (M. Marcus et al., 1993). For TinyLSTM, a sentence's plausibility score is estimated as the average surprisal (Hale, 2001; Levy, 2008) of each sentence token $w_i$ in the sequence, conditioned on the preceding sentence tokens $w_{<i}$, i.e., its conditional negative log probability.

$$surprisal(s = w_1 \dots w_n) = -1/n \sum_{i=1}^{n} log\, P(w_i \mid w_{<i})$$

The model is available through the LM Zoo library (Gauthier et al., 2020).

**Thematic fit** models the degree of semantic compatibility between an event's "prototype" verb argument, calculated from distributional text information (McRae et al., 1998), the role filler proposed by the sentence. Different extractional models to measure thematic fit have been proposed (e.g., Erk, 2007; Lenci, 2011; Greenberg et al., 2015; Sayeed et al., 2016). Here, we follow Lenci (2011) for calculating prototypical representations: Given an event, described by the predicate and subject, 1) we use Local Mutual Information (Evert, 2008) to retrieve the 200 entities most strongly associated with each (in the specific syntactic position); 2) next, we compute the intersection of the two entity lists to find the entities compatible with the compositional event description. In case the intersection is empty, we prioritize the entities associated with the verb and use only them to create the prototype; 3) we rank entities based on the product of their association scores with the subject and the verb, and select the 20 entities most strongly associated with both; 4) we compute the prototype vector as the centroid of these entities' representations, i.e., as the average of their FastText (Bojanowski et al., 2017) word embeddings. After computing the prototype representation, we obtain a sentence's plausibility score as the cosine similarity between the FastText embedding of the proposed object and the relevant prototype vector.

$$thematicFit(sentence) = cosine(\overrightarrow{w_{patient}}, \overrightarrow{w_{patient-prototype}})$$

The **Structured Distributional Model** (SDM; Chersoni et al., 2019) improves on standard models of thematic fit by leveraging insights from Discourse Representation Theory (DRT) (Kamp, 1981; Heim, 1982), a formal theory of dynamic semantics, in addition to distributional information extracted from text corpora. DRT assumes that each clause describes an event or situation, and that listeners dynamically build representations of these events as the sentence unfolds over time. The novel contribution of SDM is to infuse these dynamic discourse representations with distributional knowledge about events and their typical participants. In computing the compatibility between a proposed role filler and its distributional prototype, SDM combines two tiers of semantic meaning representation. On the one hand, SDM computes a *context-independent* representation of the linguistic context (linguistic condition; LC) via summing the embeddings associated with all lexical items in the leftward context. On the other hand, SDM computes a *context-dependent* representation of the prototypical argument via a distributional event graph (DEG) that is external to the model and was extracted from parsed text corpora. In this graph, the nodes represent the lexical items in the corpus and the edges encode the statistical syntactic relations between these items. Given a set of linguistic items, SDM queries the DEG for the most common role fillers associated with the items in the active context (AC) and computes a prototype representation for that slot as the centroid of FastText embeddings from the highest-ranked entities. A sentence's plausibility score is finally calculated as the sum of the average cosine distance between the representations of each proposed verb argument filler $w_i$ (provided by the linguistic input) with (i) the average representation of the preceding context, $LC(sentence_{<i})$) and (ii) the context-dependent prototype for the target role, $AC(sentence_{<i})$.

$$SDM(sentence) = \sum_{i \in \{agent, patient\}} \frac{cosine(\overrightarrow{w_i}, LC(sentence_{<i})) + cosine(\overrightarrow{w_i}, AC(sentence_{<i}))}{2}$$

Finally, the syntax-based PPMI (**PPMI-syntax)** model is trained to encode statistical associations between verbs and their dependents, on a concatenation of the dependency-parsed ukWaC corpus (Baroni et al., 2009), a dump of the English Wikipedia from 2018, and the British National Corpus (Leech, 1992). For each sentence, we first extract triplets of syntactic relations <verbal head, nominal dependent, syntactic role> of minimum frequency 2, and compute the Positive

Pointwise Mutual Information (PPMI) score for each such triplet (with N = total frequency of all triplets):

$$PPMI(head, dependent, relation) = max(0, \frac{f(head,dependent,relation) * N}{f(head,*,relation)\ f(*,dependent,relation)})$$

In the testing phase, a sentence's plausibility score is then computed as the sum of the PPMI score for the verb and the subject, and the PPMI score for the verb and the object. We apply Laplace smoothing (also called add-one smoothing) consisting of adding 1 to all the counts.

# SI3. Baseline models: detailed results (Dataset 1)

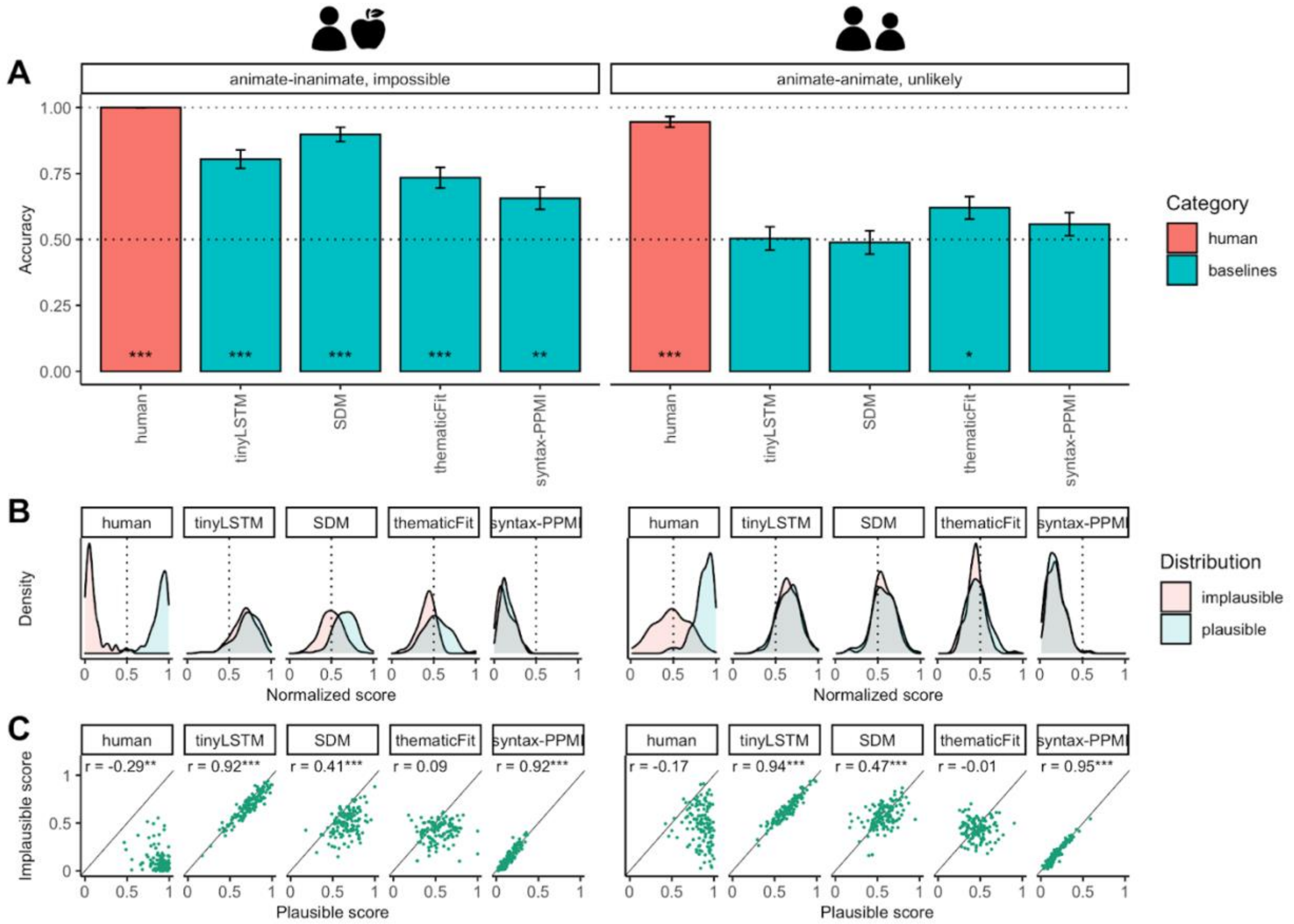

***Figure S1**. Baseline performance on Dataset 1 (results in A are the same as in Figure 1 in the main text). **(A)** Human and baseline model accuracy scores for AI and AA sentence pairs. (**B)** Density plots for plausible and implausible sentences. The dotted line shows the midpoint on the normalized score scale (0.5). **(C)** Correlation plots for plausible and implausible sentences. Each dot represents a sentence score. The diagonal is an identity line. Annotations show Pearson r correlation values and significance levels.*

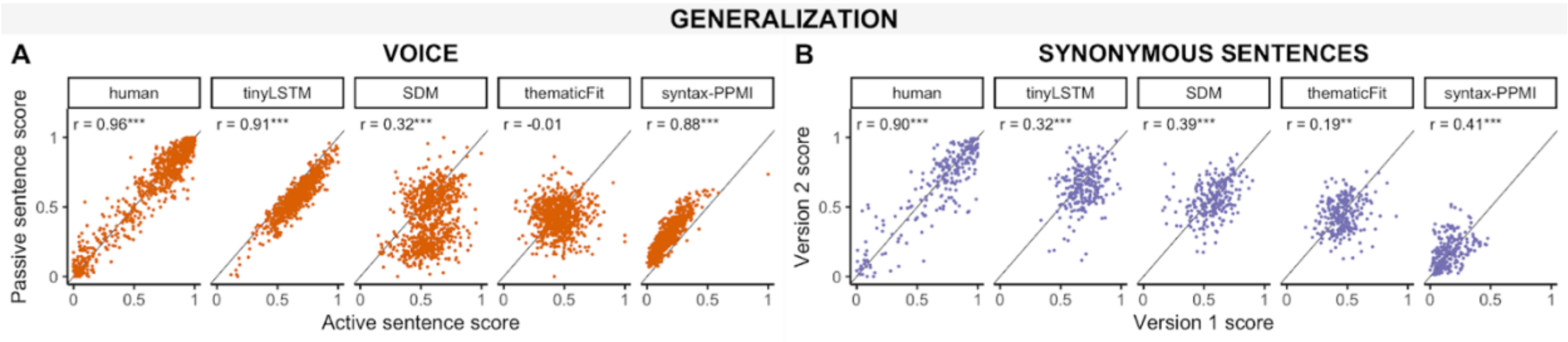

***Figure S2***. *Baseline model generalization performance on Dataset 1.* ***(A)*** *Human and baseline model scores for active voice and passive voice versions of the same sentence.* ***(B)*** *Human and baseline model scores for synonymous sentences. Each dot represents a sentence score. The diagonal is an identity line. Correlation values (Pearson's r) show correlation between sentence pairs.*

***Table S2***. *Mixed effects modeling results. Effects that are significant in humans are highlighted in bold.*

| | | humans | tinyLSTM | SDM | thematicFit | syntax-PPMI | Mean across models |
|---|---|---|---|---|---|---|---|
| **Core effects** | **Implausible AA > Plausible AA** | **-0.38 ***** | | | **-0.04 *** | | **-0.01** |
| | **Implausible AI > Implausible AA** | **-0.37 ***** | **-0.04 ***** | | **-0.04 *** | | **-0.02** |
| Surface-level effects | Voice (active>passive) | | -0.18 *** | 0.37 *** | 0.06 * | -0.26 *** | 0 |
| | Agent frequency | | 0.02 *** | 0.03 *** | | -0.03 *** | 0.01 |
| | Patient frequency | | 0.03 *** | 0.04 *** | | -0.03 *** | 0.01 |
| | Verb frequency | | 0.02 * | 0.04 *** | | -0.04 *** | 0 |
| | Avg. word frequency | | | | | | 0 |
| | Sentence length | | -0.13 *** | 0.12 *** | | -0.07 *** | -0.02 |
| | Voice x Sentence (AA>control) | | | 0.06 * | | | 0.01 |
| | **Voice x Sentence (AI>AA)** | **0.03 **** | **0.01 *** | **0.13 ***** | **0.12 ***** | | **0.07** |
| | Plausibility x Voice x Sentence (AA>control) | | | | | | 0 |
| | **Plausibility x Voice x Sentence (AI>AA)** | **-0.07 ***** | | **-0.27 ***** | **-0.11 ***** | | **-0.1** |

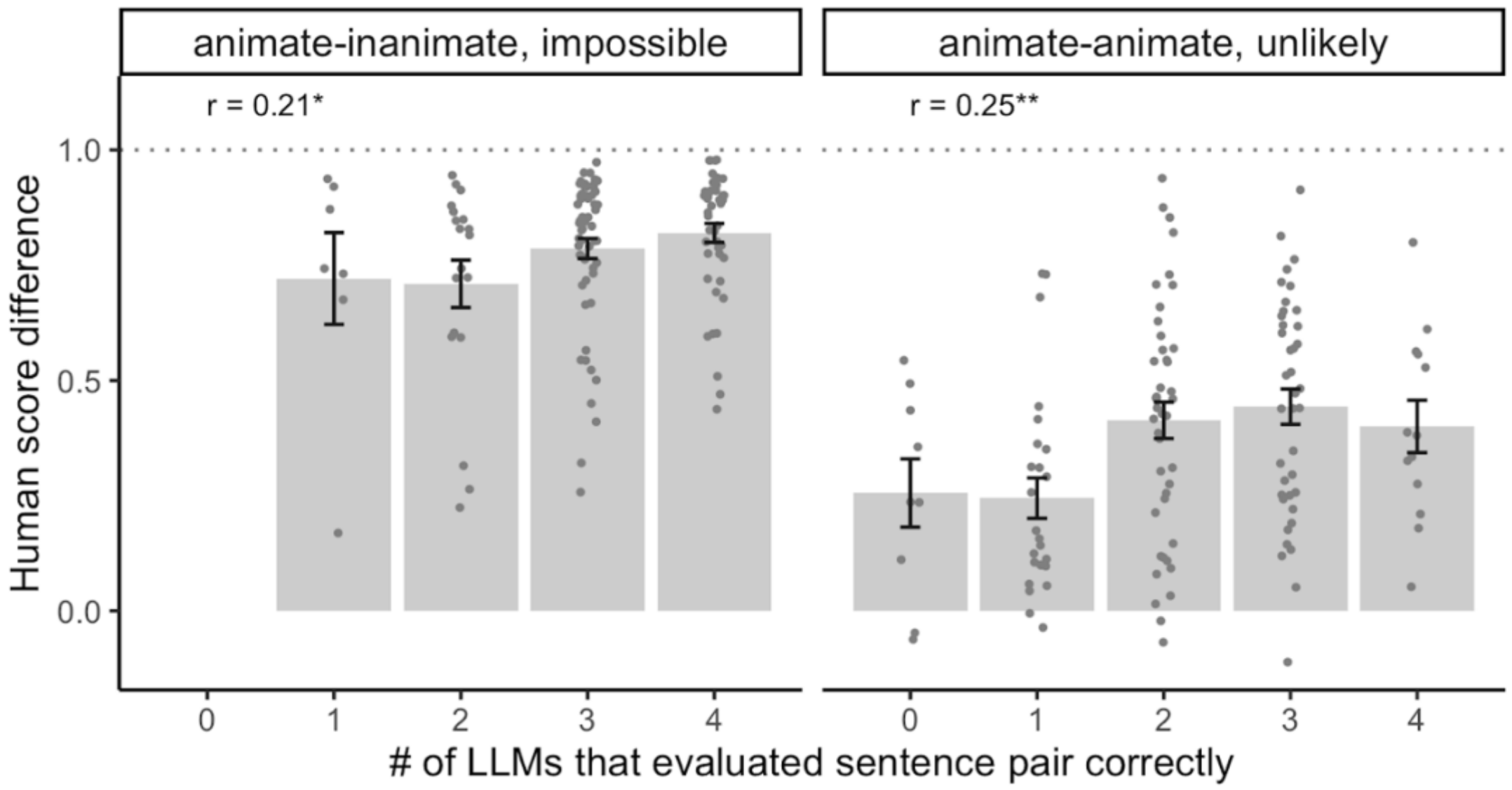

***Figure S3**. Baseline model error analysis, Dataset 1. There are no sentence pairs where the human score difference was positive that all baseline models got wrong, a result consistent with the high heterogeneity of the baselines. However, the positive correlation between human score differences and the number of correct model responses does indicate that sentence pairs with more polarized scores are easier to distinguish using distributional linguistic features.*

***Table S3.** Sentences that were evaluated incorrectly by all or most (3 out of 4) baseline models, ordered by human score difference in descending order. Sentences where the human ratings also deviated from the ground truth labels are grayed out.*

| | **Trial type** | **#LLMs correct (of 4)** | **Human score difference** | **Plausible sentence** | **Implausible sentence** |
|---|---|---|---|---|---|
| 1 | AI | 1 | 0.94 | The secretary organized the desk. | The desk organized the secretary. |
| 2 | AI | 1 | 0.92 | The cook grilled the octopus. | The octopus grilled the cook. |
| 3 | AI | 1 | 0.87 | The meat-eater devoured the filet. | The filet devoured the meat-eater. |
| 4 | AI | 1 | 0.74 | The journalist ditched the article. | The article ditched the journalist. |
| 5 | AA | 1 | 0.73 | The artisan trained the apprentice. | The apprentice trained the artisan. |
| 6 | AA | 1 | 0.73 | The chauffeur drove the diplomat. | The diplomat drove the chauffeur. |
| 7 | AI | 1 | 0.73 | The operative blew the assignment. | The assignment blew the operative. |
| 8 | AA | 1 | 0.68 | The masseuse relaxed the linebacker. | The linebacker relaxed the masseuse. |
| 9 | AI | 1 | 0.68 | The nutritionist detested the marmalade. | The marmalade detested the nutritionist. |
| 10 | AA | 0 | 0.54 | The lion chased the tour-guide. | The tour-guide chased the lion. |
| 11 | AA | 0 | 0.49 | The brunette tipped the busboy. | The busboy tipped the brunette. |

| | | | | |
|---|---|---|---|---|
| 12 AA | 0 | 0.44 | The dressmaker attired the ballerina. | The ballerina attired the dressmaker. |
| 13 AA | 1 | 0.44 | The barber shaved the old man. | The old man shaved the barber. |
| 14 AA | 1 | 0.42 | The king exiled the rebel. | The rebel exiled the king. |
| 15 AA | 0 | 0.36 | The pessimist discouraged the contestant. | The contestant discouraged the pessimist. |
| 16 AA | 1 | 0.36 | The terrorist petrified the first lady. | The first lady petrified the terrorist. |
| 17 AA | 1 | 0.35 | The policeman subdued the rabble-rouser. | The rabble-rouser subdued the policeman. |
| 18 AA | 1 | 0.31 | The monarch banished the insurgent. | The insurgent banished the monarch. |
| 19 AA | 1 | 0.31 | The miscreant kidnapped the beneficiary. | The beneficiary kidnapped the miscreant. |
| 20 AA | 1 | 0.29 | The alcoholic hassled the guest. | The guest hassled the alcoholic. |
| 21 AA | 1 | 0.26 | The impersonator conned the inspector. | The inspector conned the impersonator. |
| 22 AA | 0 | 0.24 | The tennis player thanked the chiropractor. | The chiropractor thanked the tennis player. |
| 23 AA | 0 | 0.24 | The TV station head promoted the newsagent. | The newsagent promoted the TV station head. |
| 24 AA | 1 | 0.17 | The entrepreneur hired the specialist. | The specialist hired the entrepreneur. |
| 25 AI | 1 | 0.17 | The hatter decorated the bowler. | The bowler decorated the hatter. |
| 26 AA | 1 | 0.16 | The environmentalist cautioned the tobacconist. | The tobacconist cautioned the environmentalist. |
| 27 AA | 1 | 0.14 | The playboy courted the damsel. | The damsel courted the playboy. |
| 28 AA | 1 | 0.12 | The nurse helped the orthodontist. | The orthodontist helped the nurse. |
| 29 AA | 0 | 0.11 | The nomad cherished the clergyman. | The clergyman cherished the nomad. |
| 30 AA | 1 | 0.11 | The drunk bothered the visitor. | The visitor bothered the drunk. |
| 31 AA | 1 | 0.11 | The streetwalker undercharged the seaman. | The seaman undercharged the streetwalker. |
| 32 AA | 1 | 0.1 | The genius shocked the cousin. | The cousin shocked the genius. |
| 33 AA | 1 | 0.1 | The windbag taunted the recluse. | The recluse taunted the windbag. |
| 34 AA | 1 | 0.1 | The judge praised the gold medalist. | The gold medalist praised the judge. |
| 35 AA | 1 | 0.06 | The prodigy surprised the relative. | The relative surprised the prodigy. |
| 36 AA | 1 | 0.05 | The extortionist menaced the legislator. | The legislator menaced the extortionist. |
| 37 AA | 1 | 0.04 | The arsonist alarmed the vendor. | The vendor alarmed the arsonist. |
| *38 AA* | *1* | *-0.01* | *The liar emulated the victor.* | *The victor emulated the liar.* |
| *39 AA* | *1* | *-0.04* | *The reviewer criticized the right-winger.* | *The right-winger criticized the reviewer.* |
| *40 AA* | *0* | *-0.05* | *The admirer badgered the director.* | *The director badgered the admirer.* |
| *41 AA* | *0* | *-0.06* | *The critic attacked the conservative.* | *The conservative attacked the critic.* |

## SI4. Dataset 1 additional results

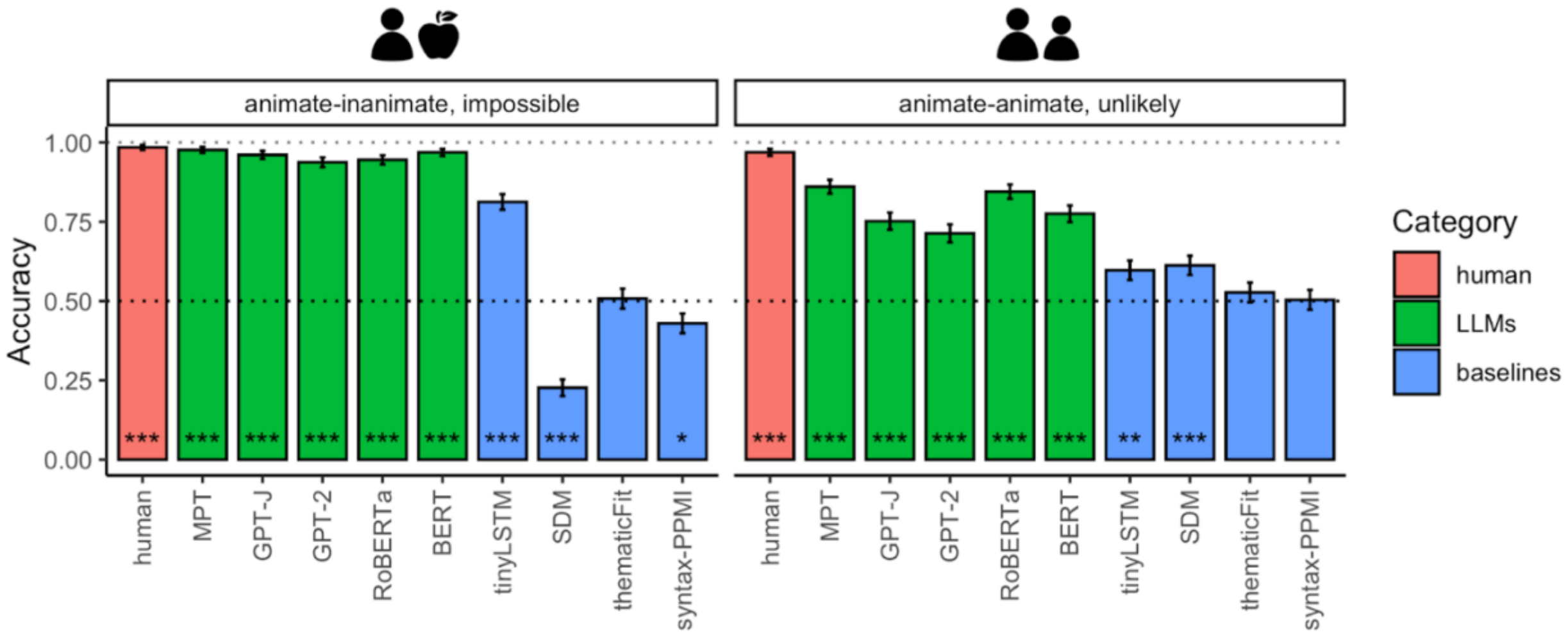

***Figure S4***. *Binary accuracy results on passive sentence versions from Dataset 1.*

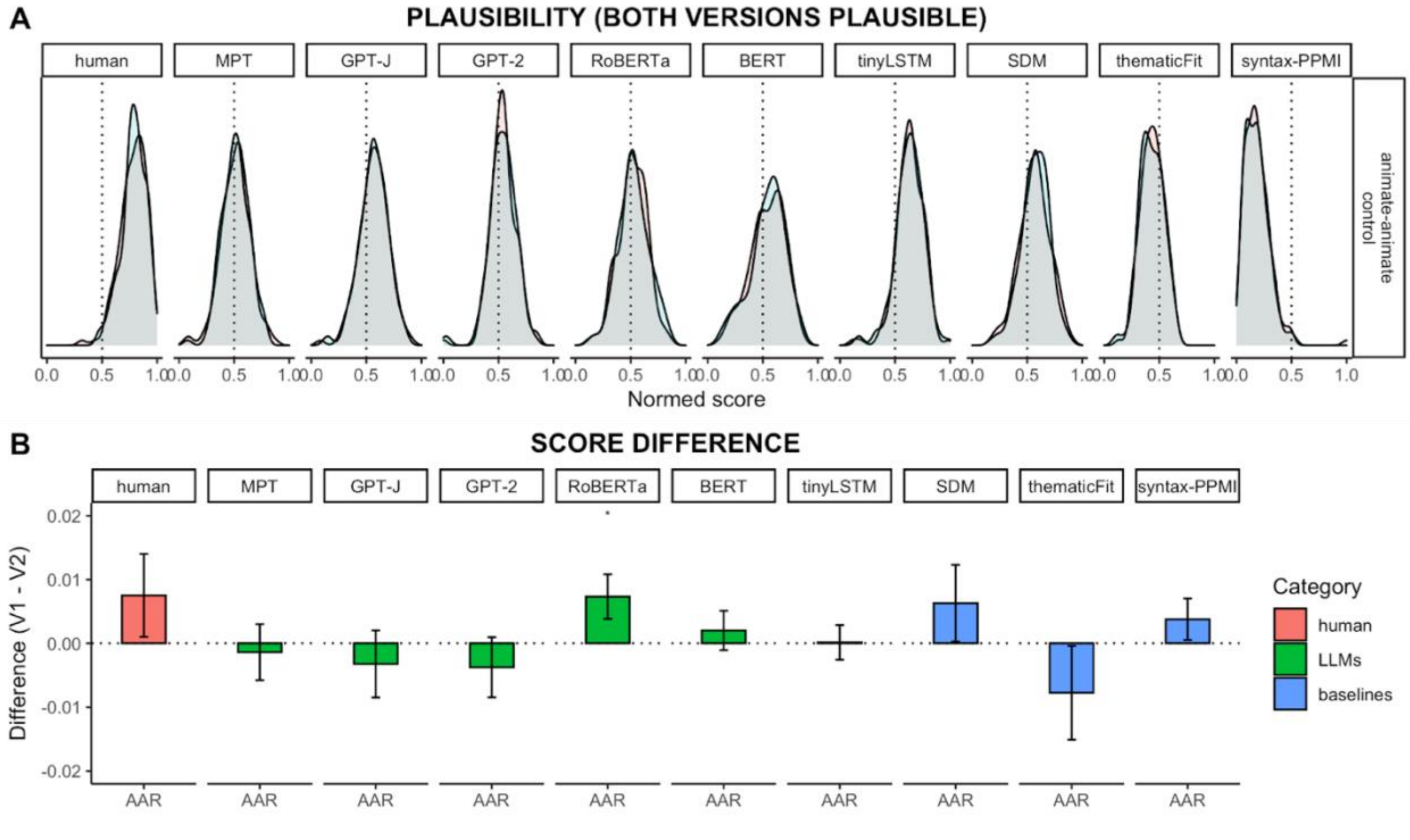

***Figure S5***. *Human, LLM and baseline model plausibility score distributions (A) and score differences (B) for AA-control sentences, i.e., those where both versions are plausible. Error bars show the standard error of difference in scores across sentence pairs.*

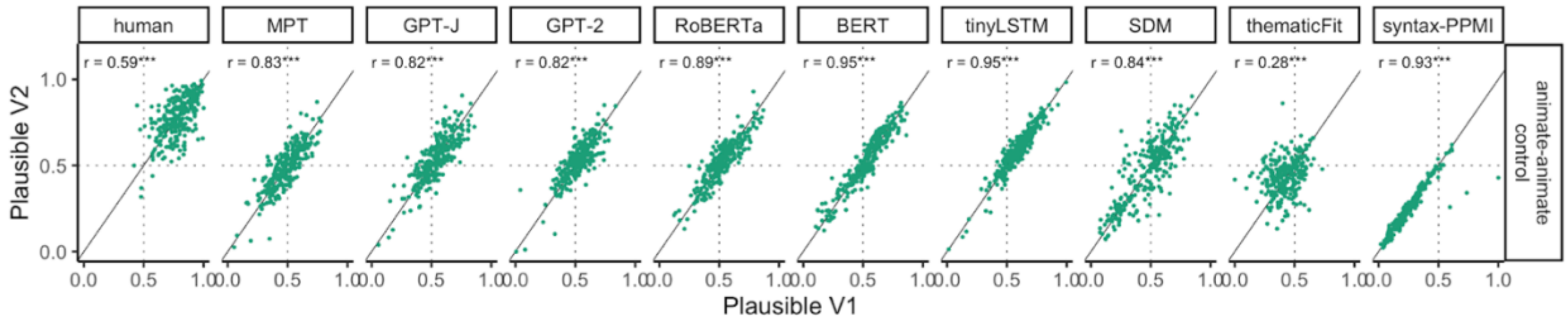

***Figure S6***. *Correlation between plausibility scores for AA-control sentences (both versions plausible). Each dot represents a sentence score. The diagonal is an identity line. The dotted lines show the midpoint on the normalized score scale (0.5). Both human and model scores show a positive correlation, indicating that word-level properties influence the ratings; however, only human scores are consistently located in the upper-right corner of the plot ("both versions plausible").*

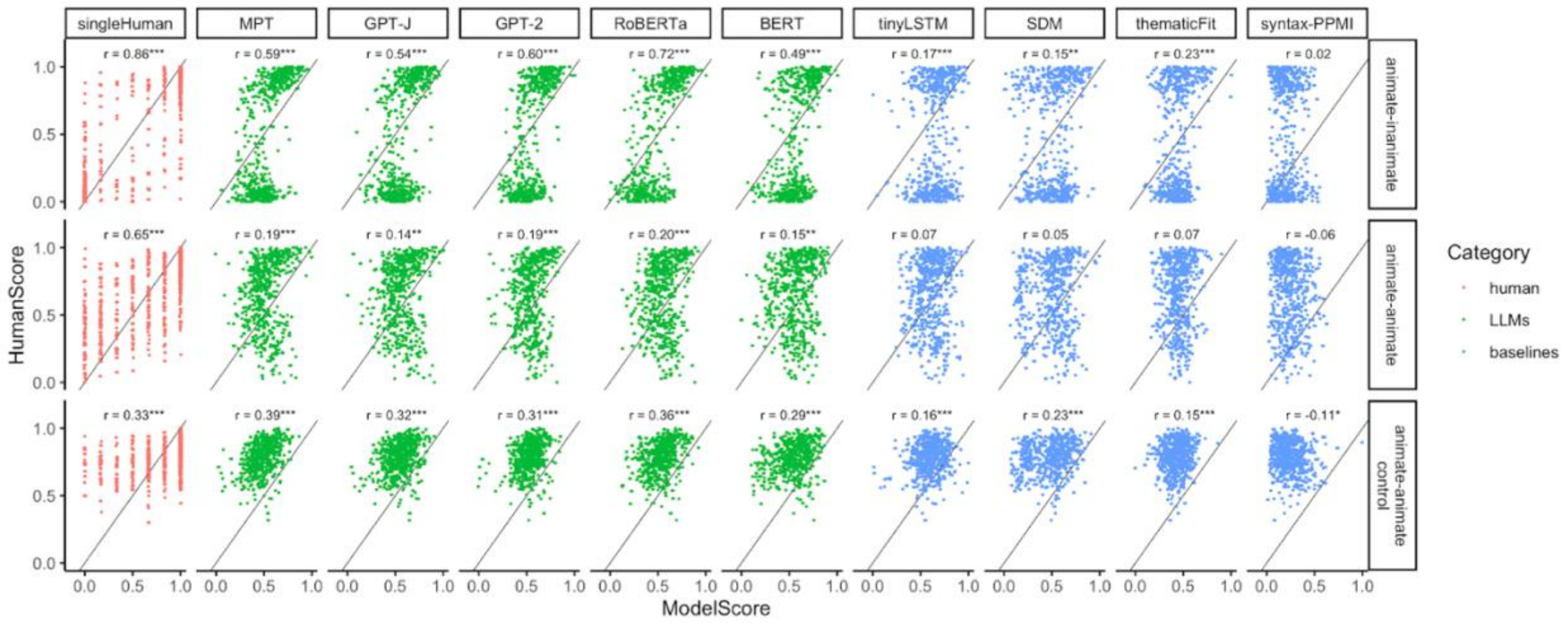

***Figure S7***. *Correlation between the average human plausibility ratings and model scores for AI (top), AA (middle), and AA-control (bottom) sentences. Each dot represents a sentence score. The diagonal is an identity line. Left column: the correlation between a randomly selected score from a single participant for each sentence and the average of the remaining human participant scores for that sentence.*

***Table S4***. *Dataset 1 AA sentences where the human judgments deviated from the ground truth labels, ordered by human score difference in descending order.*

| | Trial type | Human score difference | Sentence labeled as plausible | Sentence labeled as implausible |
|---|---|---|---|---|
| 1 | AA | -0.01 | The liar emulated the victor. | The victor emulated the liar. |
| 2 | AA | -0.02 | The vocalist disillusioned the connoisseur. | The connoisseur disillusioned the vocalist. |
| 3 | AA | -0.04 | The reviewer criticized the right-winger. | The right-winger criticized the reviewer. |
| 4 | AA | -0.05 | The admirer badgered the director. | The director badgered the admirer. |

| | | | |
|---|---|---|---|
| 5 AA | -0.06 | The critic attacked the conservative. | The conservative attacked the critic. |
| 6 AA | -0.07 | The pixie mesmerized the ogre. | The ogre mesmerized the pixie. |
| 7 AA | -0.11 | The orderly assisted the dentist. | The dentist assisted the orderly. |

# SI5 Scaling results

We investigate the performance of GPT or GPT-style, decoder-only Transformer language models on our datasets, across different model sized:

***Table S5**. Overview of model sizes.*

| Model | Number of parameters |
|---|---|
| **DistilGPT-2** | 82M |
| **GPT-2** | 110M |
| **GPT-2-medium** | 345M |
| **GPT-2-large** | 774M |
| **GPT-2-xl** | 1.6B |
| **GPT-J-6b** | 6B |
| **MPT-30b** | 30B |

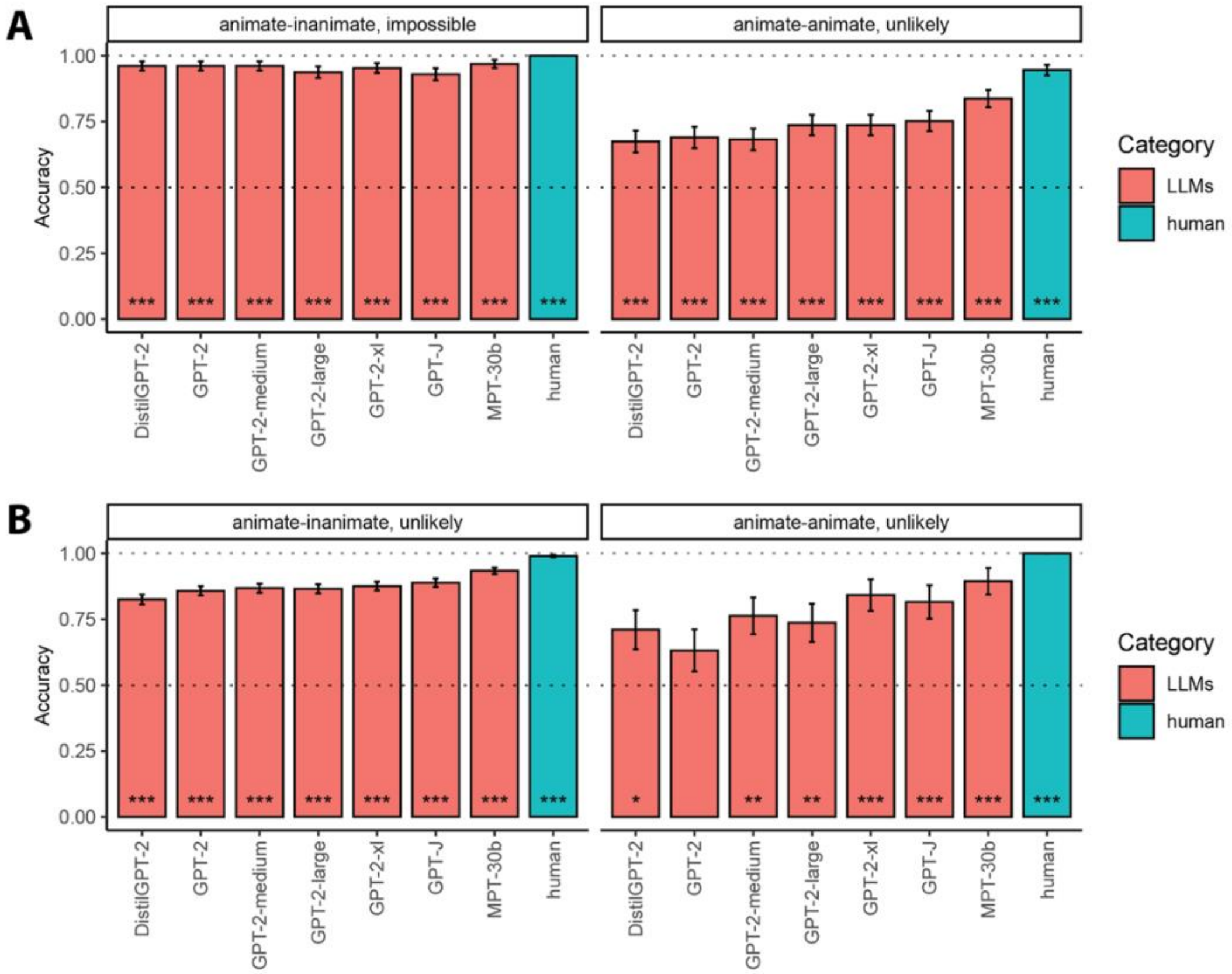

***Figure S8***. *Comparison of the effect of model size on binary accuracy performance within unidirectional models.* ***(A)*** *Results for Dataset 1.* ***(B)*** *Results for Dataset 2 (left) and Dataset 3 (right). Models are arranged from smallest (left) to largest (right).*

# SI6. Datasets 2 and 3, detailed results

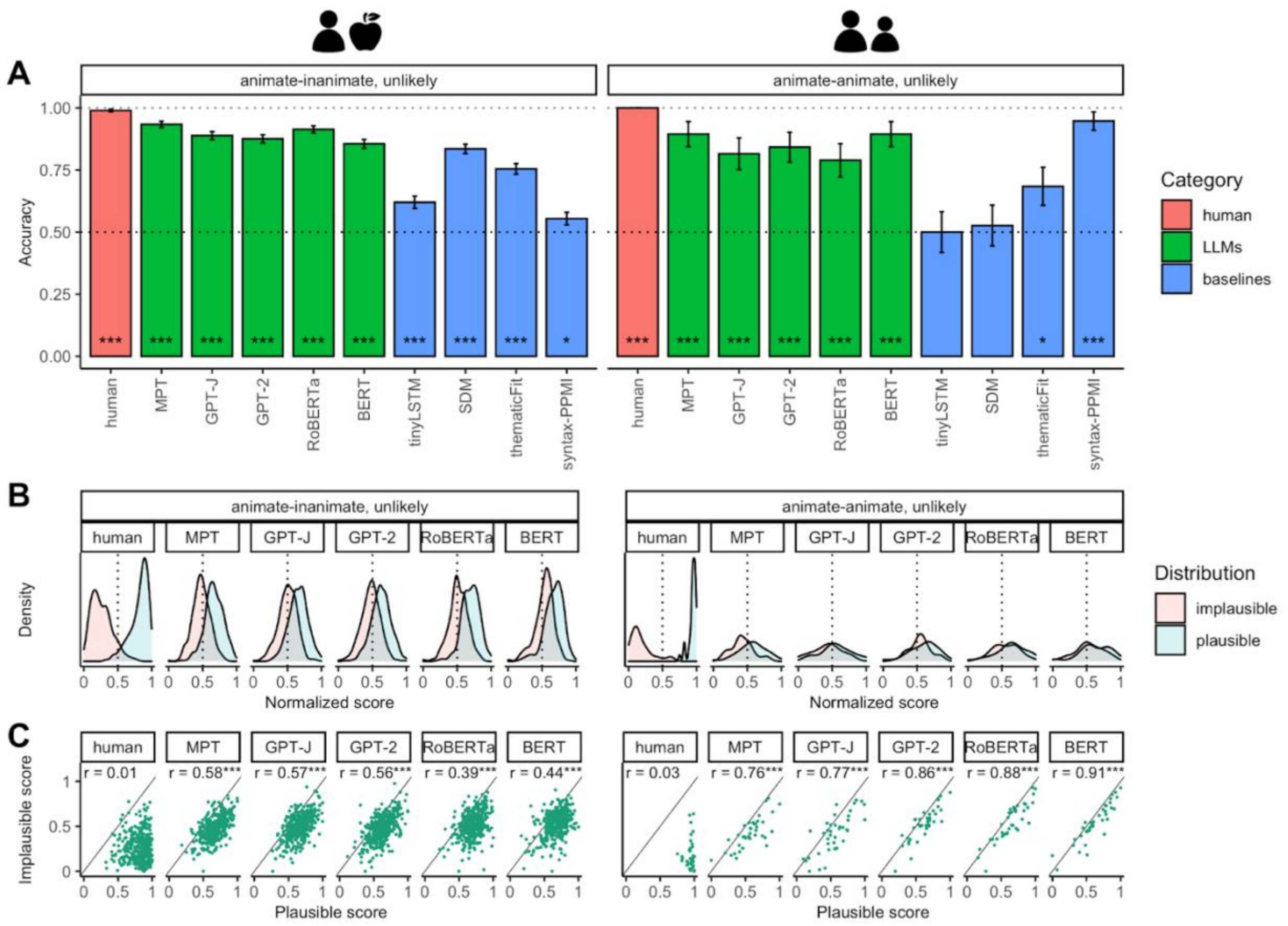

***Figure S9***. *Model performance on Dataset 2 (left) and Dataset 3 (right).* ***(A)*** *Human and baseline model accuracy scores. (****B)*** *Density plots for plausible and implausible sentences. The dotted line shows the midpoint on the normalized score scale (0.5).* ***(C)*** *Human and LLM scores for active voice and passive voice versions of the same sentence.* ***(D)*** *Human and LLM scores for synonymous sentences. In C each dot represents a sentence score. The diagonal is an identity line.*

***Table S6***. *Mixed effects modeling results for Dataset 2.*

| | | humans | MPT | GPT-J | GPT-2 | RoBERTa | BERT | Mean across models |
|---|---|---|---|---|---|---|---|---|
| **Core effects** | **Plausibility** | **-0.55 ***** | **-0.18 ***** | **-0.15 ***** | **-0.14 ***** | **-0.18 ***** | **-0.13 ***** | **-0.16** |
| Surface-level effects | Agent frequency | | | | | | | 0.01 |
| | Patient frequency | | | | | | | 0.01 |
| | Verb frequency | | | | | | | -0.01 |
| | Avg. word frequency | | | | 0.04 * | | | 0.02 |
| | Sentence length | | | -0.04 *** | -0.04 *** | -0.06 *** | -0.09 *** | -0.05 |

***Table S7**. Mixed effects modeling results for Dataset 3.*

| | | humans | MPT | GPT-J | GPT-2 | RoBERTa | BERT | Mean across models |
|---|---|---|---|---|---|---|---|---|
| **Core effects** | **Plausibility** | **-0.76 ***** | **-0.15 ***** | **-0.14 ***** | **-0.1 ***** | **-0.1 ***** | **-0.09 ***** | **-0.12** |
| Surface-level effects | Agent frequency | **0.05 *** | | **0.08 *** | **0.08 *** | **0.12 **** | **0.08 *** | **0.08** |
| | Patient frequency | | 0.08 * | | 0.08 * | 0.12 *** | 0.07 * | 0.08 |
| | Verb frequency | | | | | | | 0.04 |
| | Avg. word frequency | | | | | | | 0 |
| | Sentence length | | -0.04 ** | -0.06 ** | -0.05 ** | | -0.07 *** | -0.04 |

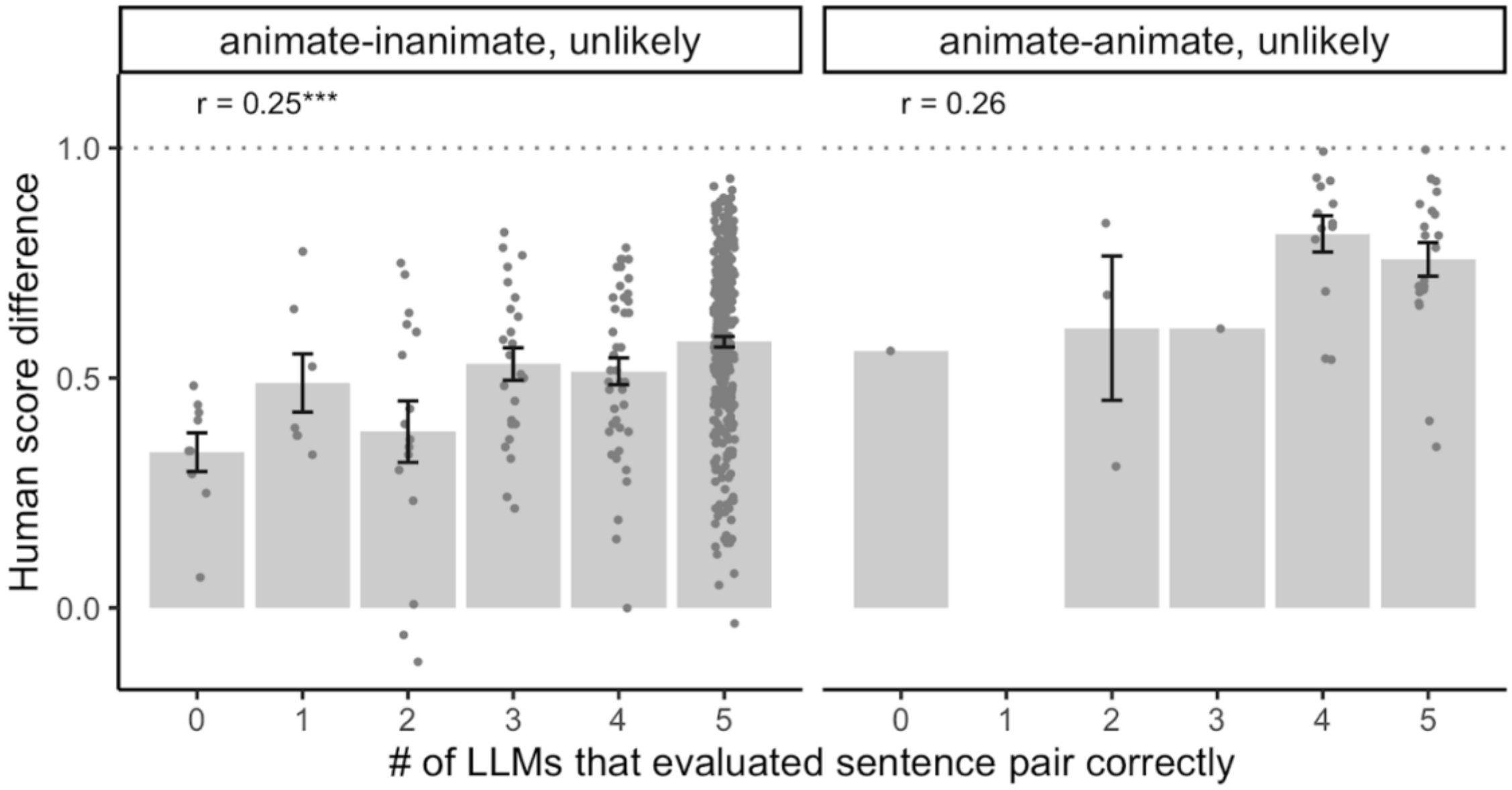

***Figure S10**. LLM error analysis, Datasets 2 (left) and 3 (right). Each dot is a sentence; error bars denote standard errors of the mean.*

***Table S8.*** *Sentences from Dataset 2 that were evaluated incorrectly by all LLMs, ordered by human score difference in descending order.*

| | #LLMs correct (of 5) | Human score difference | Plausible sentence | Implausible sentence |
|---|---|---|---|---|
| 1 | 0 | 0.48 | The child wrote the conjugation. | The child wrote the diagnosis. |
| 2 | 0 | 0.44 | The child drank the coke. | The child drank the beer. |
| 3 | 0 | 0.42 | The woman painted the toenail. | The woman painted the sign. |
| 4 | 0 | 0.41 | The ant stacked the supply. | The ant stacked the suitcase. |
| 5 | 0 | 0.34 | The child crossed the park. | The child crossed the river. |
| 6 | 0 | 0.34 | The policeman hit the demonstrator. | The policeman hit the ball. |
| 7 | 0 | 0.29 | The guest held the drink. | The guest held the camera. |
| 8 | 0 | 0.25 | The porter stacked the suitcase. | The porter stacked the wood. |
| 9 | 0 | 0.07 | The magician read the hand. | The magician read the newspaper. |

***Table S9****. Sentences from Dataset 3 that were evaluated incorrectly by half of or more of the LLMs, ordered by human score difference in descending order.*

| | #LLMs correct (of 5) | Human score difference | Plausible sentence | Implausible sentence |
|---|---|---|---|---|
| 1 | 2 | 0.84 | The professor is lecturing to the student. | The student is lecturing to the professor. |
| 2 | 2 | 0.68 | The lawyer is giving money to the beggar. | The beggar is giving money to the lawyer. |
| 3 | 0 | 0.56 | The visitor is pushing the patient. | The patient is pushing the visitor. |
| 4 | 2 | 0.31 | The father is tying the son's shoes. | The son is tying the father's shoes. |

# SI7. Alternative metrics for LLM models

For bidirectional models, which cannot evaluate the likelihood of a sentence via chain rule, it is unclear whether a sentence's pseudo-log-likelihood score (Kauf & Ivanova, 2023; Salazar et al., 2020) is a good proxy for sentence plausibility. To investigate the effect of metric choice for evaluating the plausibility of an event, we additionally compare the following ways of computing sentence plausibility under a bidirectional model:

- Last-word probability, i.e., the average log-likelihood of the subtokens that compose the last word in the sequence according to the model's tokenizer

  - $P(s) = 1/(t' - t)\sum_{i=t}^{t'} \log P(w_i \mid w_1 \ldots w_{n-t})$

- Verb probability, i.e., the average log-likelihood of the verb's tokens $v = w_t \ldots w_{t'}$ conditioned on their bidirectional sentence context

  - $P(s) = 1/(t' - t)\sum_{i=t}^{t'} \log P(w_i \mid w_1 \ldots w_{i-1}, w_{i+1} \ldots w_n)$

- Left-to-right (l2r) causal sentence-generation probability, i.e., average log-likelihood for each token $w_i$ in the sequence, conditioned on only the preceding tokens $w_{<i}$ according to the model.

We find that a sentence's pseudo-log-likelihood score is a more robust indicator of event knowledge in bidirectional LLMs than other prediction-based metrics, such as last-word- or verb-production likelihood (**Figures S11** and **S12**). This finding aligns with recent research showing that estimating the plausibility of a proposition via comparison of the prediction probabilities for target linguistic items at a single masked-out position (as used in e.g., Kocijan et al., 2019; Abdou et al., 2020; Pedinotti et al., 2021) can result in underestimation of model performance. This underestimation derives in part from the number of suitable competitors (i.e., other surface forms representing the same underlying concept, such as *computer* and *PC*) across which the LLM has to split its probability mass (Holtzman et al., 2021) and biases derived from factors irrelevant to the task, such as a word's number of tokens under a given LLM tokenizer (Elazar et al., 2021b). Although our framework does not fully address these issues, we show that comparing likelihoods across minimal sentence pairs matched on many common confounding factors provides a principled way of estimating model performance even when the critical manipulation is not sentence-final.

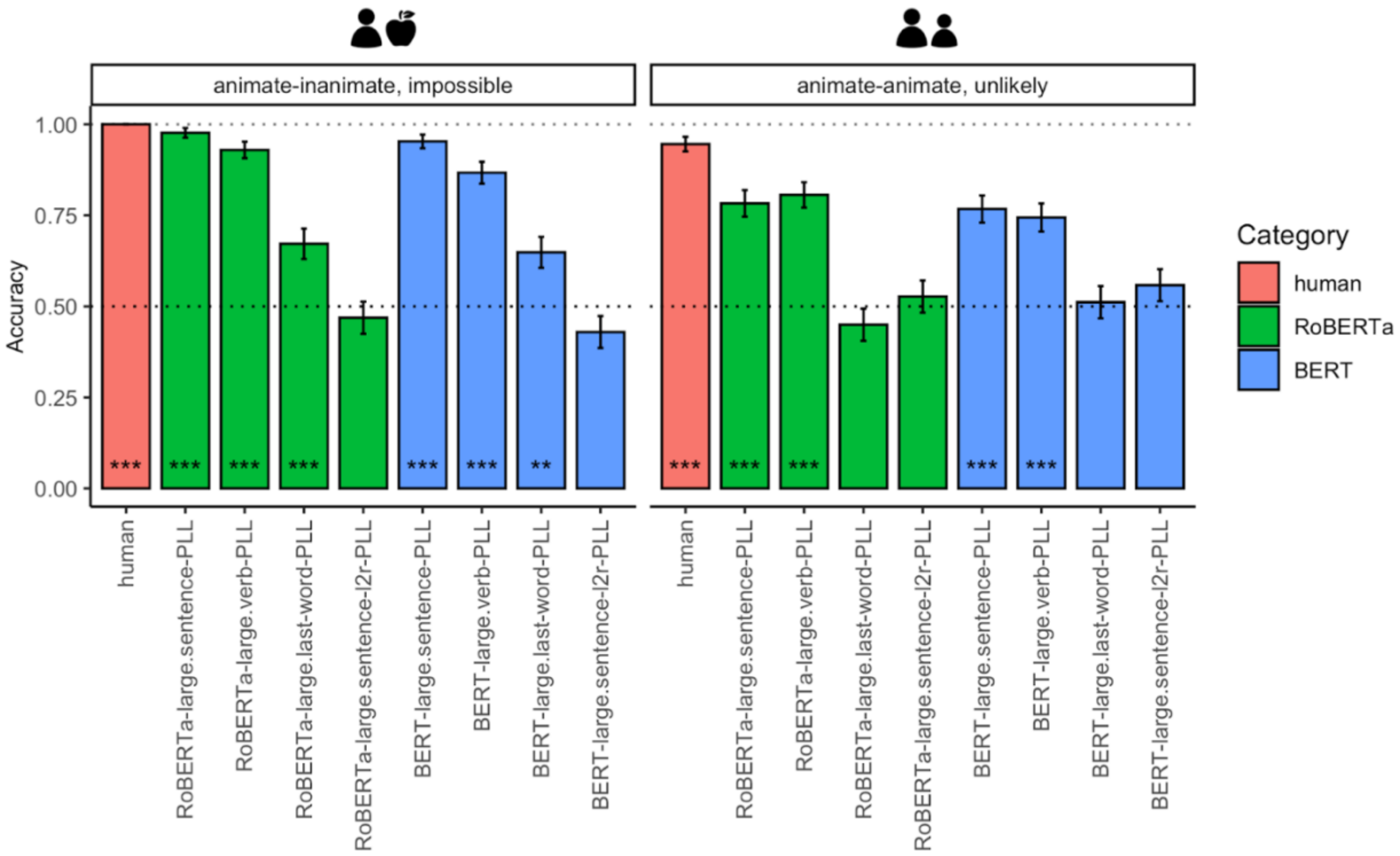

***Figure S11***. *Comparison of different metrics for bidirectional models, Dataset 1. Results that are significantly above chance (0.5) are marked with asterisks (p<0.05: *; p<0.01: **; p<0.001: ***). PLL/LL: (pseudo-)log likelihood of the sentence (used in main analyses); l2r: token-by-token sentence probability with masked right context (bidirectional models only).*

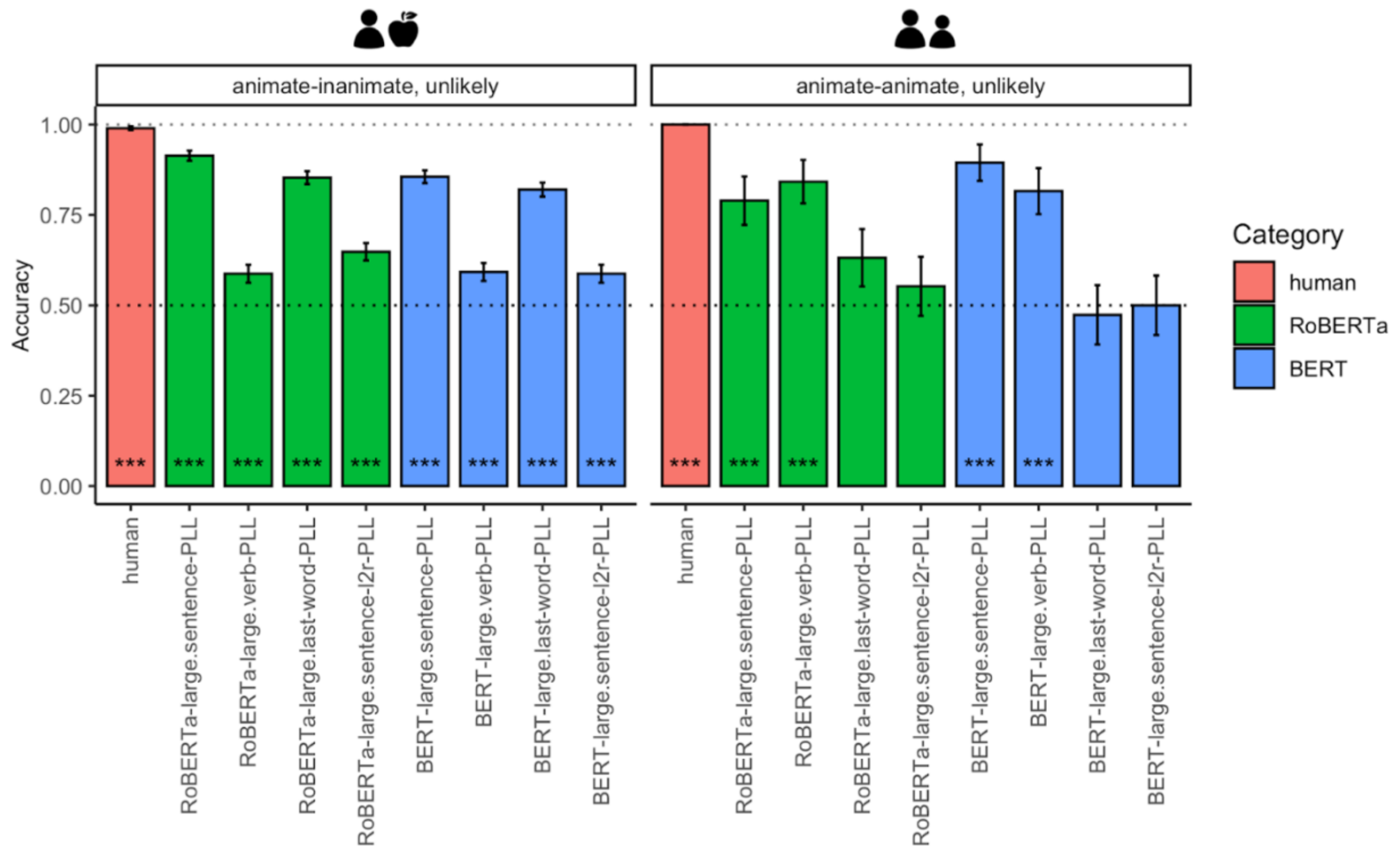

***Figure S12**. Comparison of different metrics for bidirectional models, Dataset 2 (left) and Dataset 3 (right).*

# SI8 Additional probing results

***Table S10.** Statistical analysis of probing results, generalization across trial type (AI vs AA). "Trend" refers to a linear trend within each layer group.*

| Trial Type | Parameter | MPT | GPT-J | GPT-2 | RoBERTa | BERT |
|---|---|---|---|---|---|---|
| all | Ceiling (human ratings) | 0.919 *** | 0.919 *** | 0.919 *** | 0.919 *** | 0.919 *** |
| | Early layers > human | -0.231 *** | -0.193 *** | -0.271 *** | -0.249 *** | -0.293 *** |
| | Middle layers > human | -0.053 ** | -0.067 *** | -0.109 *** | -0.11 *** | -0.261 *** |
| | Late layers > human | | -0.074 *** | -0.091 *** | -0.062 *** | -0.096 *** |
| | Early layers, trend | 0.007 *** | 0.033 *** | 0.015 *** | 0.031 *** | 0.027 *** |
| | Middle layers, trend | 0.006 *** | | 0.003 *** | 0.017 *** | 0.022 *** |
| | Late layers, trend | | | | | |
| TrainOn:AA-TestOn:AA | Ceiling (human ratings) | 0.842 *** | 0.842 *** | 0.842 *** | 0.842 *** | 0.842 *** |
| | Early layers > human | -0.275 *** | -0.234 *** | -0.278 *** | -0.282 *** | -0.326 *** |
| | Middle layers > human | -0.067 ** | -0.102 *** | -0.123 *** | -0.173 *** | -0.284 *** |
| | Late layers > human | -0.052 * | -0.102 *** | -0.1 *** | -0.077 *** | -0.124 *** |
| | Early layers, trend | 0.006 *** | 0.02 *** | 0.009 *** | 0.006 * | 0.006 * |
| | Middle layers, trend | 0.013 *** | | 0.005 *** | 0.028 *** | 0.019 *** |
| | Late layers, trend | | | | | |
| TrainOn:AA-TestOn:AI | Ceiling (human ratings) | 0.978 *** | 0.978 *** | 0.978 *** | 0.978 *** | 0.978 *** |
| | Early layers > human | -0.486 *** | -0.37 *** | -0.461 *** | -0.418 *** | -0.418 *** |
| | Middle layers > human | -0.179 *** | -0.237 *** | -0.27 *** | -0.332 *** | -0.414 *** |
| | Late layers > human | -0.151 *** | -0.253 *** | -0.264 *** | -0.153 *** | -0.311 *** |
| | Early layers, trend | 0.002 *** | 0.039 *** | 0.003 *** | 0.009 *** | 0.018 *** |
| | Middle layers, trend | 0.01 *** | -0.01 *** | 0.004 *** | 0.047 *** | 0.022 *** |
| | Late layers, trend | | 0.008 *** | | -0.011 *** | |
| TrainOn:AI-TestOn:AA | Ceiling (human ratings) | 0.731 *** | 0.731 *** | 0.731 *** | 0.731 *** | 0.731 *** |
| | Early layers > human | -0.215 *** | -0.181 *** | -0.212 *** | -0.216 *** | -0.211 *** |
| | Middle layers > human | -0.115 *** | -0.074 *** | -0.112 *** | -0.168 *** | -0.21 *** |
| | Late layers > human | -0.127 *** | -0.078 *** | -0.112 *** | -0.129 *** | -0.167 *** |
| | Early layers, trend | 0.002 *** | 0.018 *** | 0.002 *** | -0.001 * | 0.004 *** |
| | Middle layers, trend | 0.006 *** | | 0.002 *** | 0.01 *** | 0.004 *** |
| | Late layers, trend | -0.002 *** | -0.001 * | -0.001 *** | 0.006 *** | -0.002 * |
| TrainOn:AI-TestOn:AI | Ceiling (human ratings) | 0.977 *** | 0.977 *** | 0.977 *** | 0.977 *** | 0.977 *** |
| | Early layers > human | -0.117 *** | -0.081 *** | -0.159 *** | -0.167 *** | -0.246 *** |
| | Middle layers > human | | | | | -0.141 *** |
| | Late layers > human | | | | | |
| | Early layers, trend | 0.015 *** | 0.038 *** | 0.026 *** | 0.057 *** | 0.049 *** |
| | Middle layers, trend | | | | | 0.024 *** |
| | Late layers, trend | | | | | |

***Table S11.*** *Statistical analysis of probing results, generalization across voice type.*

| Voice Type | Parameter | MPT | GPT-J | GPT-2 | RoBERTa | BERT |
|---|---|---|---|---|---|---|
| all | Ceiling (human ratings) | 0.915 *** | 0.915 *** | 0.915 *** | 0.915 *** | 0.915 *** |
| | Early layers > human | -0.32 *** | -0.281 *** | -0.348 *** | -0.291 *** | -0.343 *** |
| | Middle layers > human | -0.062 *** | -0.08 *** | -0.16 *** | -0.12 *** | -0.291 *** |
| | Late layers > human | | -0.062 *** | -0.14 *** | -0.061 *** | -0.129 *** |
| | Early layers, trend | 0.007 *** | 0.031 *** | 0.009 *** | 0.028 *** | 0.017 *** |
| | Middle layers, trend | 0.011 *** | 0.005 *** | 0.005 *** | 0.018 *** | 0.016 *** |
| | Late layers, trend | | | | | |
| TrainOn:active-TestOn:active | Ceiling (human ratings) | 0.919 *** | 0.919 *** | 0.919 *** | 0.919 *** | 0.919 *** |
| | Early layers > human | -0.231 *** | -0.193 *** | -0.271 *** | -0.249 *** | -0.293 *** |
| | Middle layers > human | -0.053 ** | -0.067 *** | -0.109 *** | -0.11 *** | -0.261 *** |
| | Late layers > human | | -0.074 *** | -0.091 *** | -0.062 *** | -0.096 *** |
| | Early layers, trend | 0.007 *** | 0.033 *** | 0.015 *** | 0.031 *** | 0.027 *** |
| | Middle layers, trend | 0.006 *** | | 0.003 *** | 0.017 *** | 0.022 *** |
| | Late layers, trend | | | | | |
| TrainOn:active-TestOn:passive | Ceiling (human ratings) | 0.906 *** | 0.906 *** | 0.906 *** | 0.906 *** | 0.906 *** |
| | Early layers > human | -0.555 *** | -0.462 *** | -0.499 *** | -0.482 *** | -0.486 *** |

| Dataset | Parameter | MPT | GPT-J | GPT-2 | RoBERTa | BERT |
|---|---|---|---|---|---|---|
| | Middle layers > human | -0.222 *** | -0.338 *** | -0.349 *** | -0.299 *** | -0.445 *** |
| | Late layers > human | -0.113 *** | -0.326 *** | -0.378 *** | -0.216 *** | -0.242 *** |
| | Early layers, trend | -0.007 *** | 0.007 *** | -0.006 *** | -0.015 *** | -0.02 *** |
| | Middle layers, trend | 0.026 *** | | 0.005 *** | 0.034 *** | 0.005 * |
| | Late layers, trend | | | | -0.01 ** | 0.016 *** |
| TrainOn:passive-TestOn:active | Ceiling (human ratings) | 0.919 *** | 0.919 *** | 0.919 *** | 0.919 *** | 0.919 *** |
| | Early layers > human | -0.525 *** | -0.494 *** | -0.506 *** | -0.511 *** | -0.508 *** |
| | Middle layers > human | -0.223 *** | -0.254 *** | -0.364 *** | -0.267 *** | -0.486 *** |
| | Late layers > human | -0.134 *** | -0.193 *** | -0.323 *** | -0.284 *** | -0.284 *** |
| | Early layers, trend | -0.006 *** | 0.016 *** | -0.008 *** | -0.016 *** | -0.024 *** |
| | Middle layers, trend | 0.025 *** | 0.008 ** | 0.011 *** | 0.046 *** | |
| | Late layers, trend | | | | -0.049 *** | 0.015 *** |
| TrainOn:passive-TestOn:passive | Ceiling (human ratings) | 0.904 *** | 0.904 *** | 0.904 *** | 0.904 *** | 0.904 *** |
| | Early layers > human | -0.208 *** | -0.191 *** | -0.247 *** | -0.221 *** | -0.271 *** |
| | Middle layers > human | | -0.043 ** | -0.101 *** | -0.089 *** | -0.232 *** |
| | Late layers > human | | | -0.099 *** | | -0.108 *** |
| | Early layers, trend | 0.009 *** | 0.028 *** | 0.014 *** | 0.03 *** | 0.027 *** |
| | Middle layers, trend | 0.008 *** | 0.008 *** | | 0.012 *** | 0.016 *** |
| | Late layers, trend | | | | | |

## Probing results across the three datasets

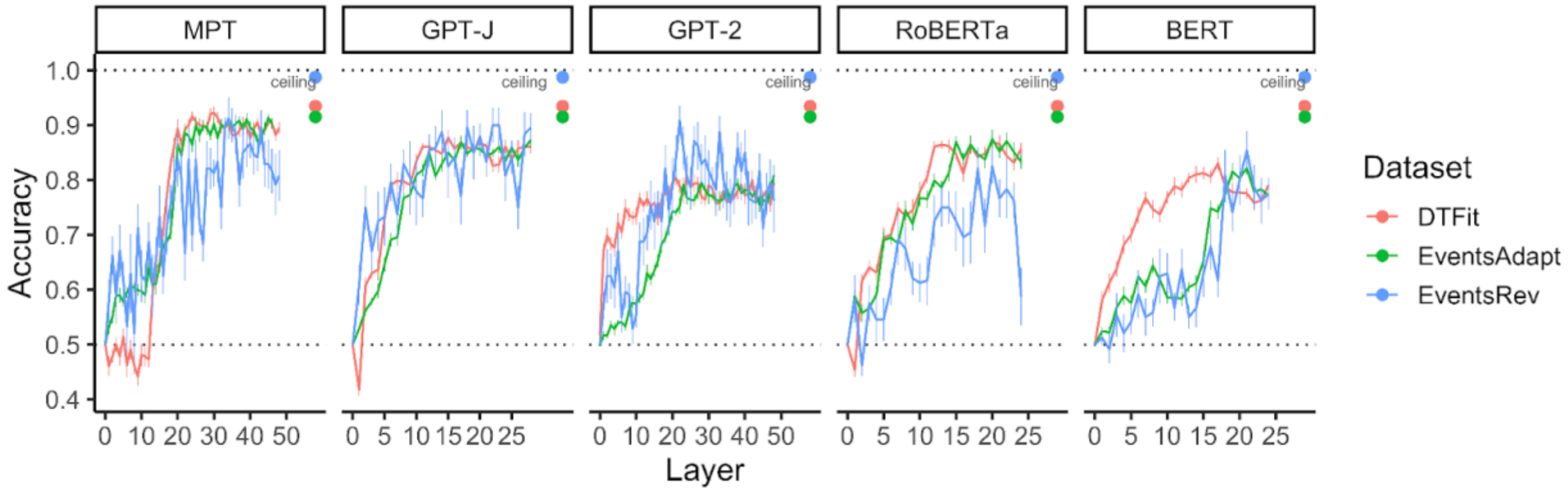

***Figure S13***. *Classification accuracies for probes trained to differentiate plausible from implausible event descriptions in model embeddings. Evaluation is separated by dataset. Ceiling values show the empirical ceiling for model performance, calculated as the classification accuracy of probes trained to differentiate plausible from implausible event descriptions based on average human judgments. Green - Dataset 1, red - Dataset 2, blue - Dataset 3.*

***Table S12***. *Statistics for probing results across the three datasets.*

| Dataset | Parameter | MPT | GPT-J | GPT-2 | RoBERTa | BERT |
|---|---|---|---|---|---|---|
| Dataset 1 | Ceiling (human ratings) | 0.92 *** | 0.92 *** | 0.92 *** | 0.92 *** | 0.92 *** |
| | Early layers > human | -0.32 *** | -0.28 *** | -0.35 *** | -0.29 *** | -0.34 *** |
| | Middle layers > human | -0.06 *** | -0.08 *** | -0.16 *** | -0.12 *** | -0.29 *** |
| | Late layers > human | | -0.06 *** | -0.14 *** | -0.06 *** | -0.13 *** |
| | Early layers, trend | 0.01 *** | 0.03 *** | 0.01 *** | 0.03 *** | 0.02 *** |
| | Middle layers, trend | 0.01 *** | 0.01 *** | 0.01 *** | 0.02 *** | 0.02 *** |
| | Late layers, trend | | | | | |
| Dataset 2 | Ceiling (human ratings) | 0.93 *** | 0.93 *** | 0.93 *** | 0.93 *** | 0.93 *** |
| | Early layers > human | -0.42 *** | -0.26 *** | -0.22 *** | -0.3 *** | -0.27 *** |
| | Middle layers > human | -0.05 ** | -0.08 *** | -0.15 *** | -0.11 *** | -0.15 *** |
| | Late layers > human | -0.04 * | -0.08 *** | -0.16 *** | -0.08 *** | -0.15 *** |
| | Early layers, trend | 0.01 *** | 0.04 *** | 0.01 *** | 0.03 *** | 0.03 *** |
| | Middle layers, trend | 0.01 *** | | | 0.01 *** | 0.01 *** |
| | Late layers, trend | | | | | |
| Dataset 3 | Ceiling (human ratings) | 0.99 *** | 0.99 *** | 0.99 *** | 0.99 *** | 0.99 *** |
| | Early layers > human | -0.37 *** | -0.27 *** | -0.36 *** | -0.41 *** | -0.45 *** |
| | Middle layers > human | -0.21 *** | -0.16 *** | -0.17 *** | -0.3 *** | -0.38 *** |
| | Late layers > human | -0.13 ** | -0.13 ** | -0.2 *** | -0.24 *** | -0.22 *** |
| | Early layers, trend | 0.01 ** | 0.03 *** | 0.01 *** | 0.02 *** | |
| | Middle layers, trend | | | | 0.02 ** | |
| | Late layers, trend | | | | | |

## Multiclass probing

In the MTurk study, humans do not perform binary classification but rather assign a graded plausibility score of 1-7 to each sentence. To investigate whether LLMs are able to predict the more fine-graded plausibility estimates, and to account for the difference between impossible and unlikely events, we trained an ordinal multiclass classifier from LLM activations to the rounded average human judgments for each sentence. Across the seven possible classes, the human score distributions are unbalanced. To enable classifier convergence, we aggregated the human estimates into three classes: “impossible”, subsuming human judgments scores of 1 and 2, “plausible”, subsuming human judgment scores of 6 and 7, and “unlikely”, subsuming human judgment scores of 3 to 5. We used sklearn’s (Pedregosa et al., 2011) Support Vector Classification module with a linear kernel and balanced class weights. We found that the overall results were similar to the results reported in the main text, with some generalization differences.

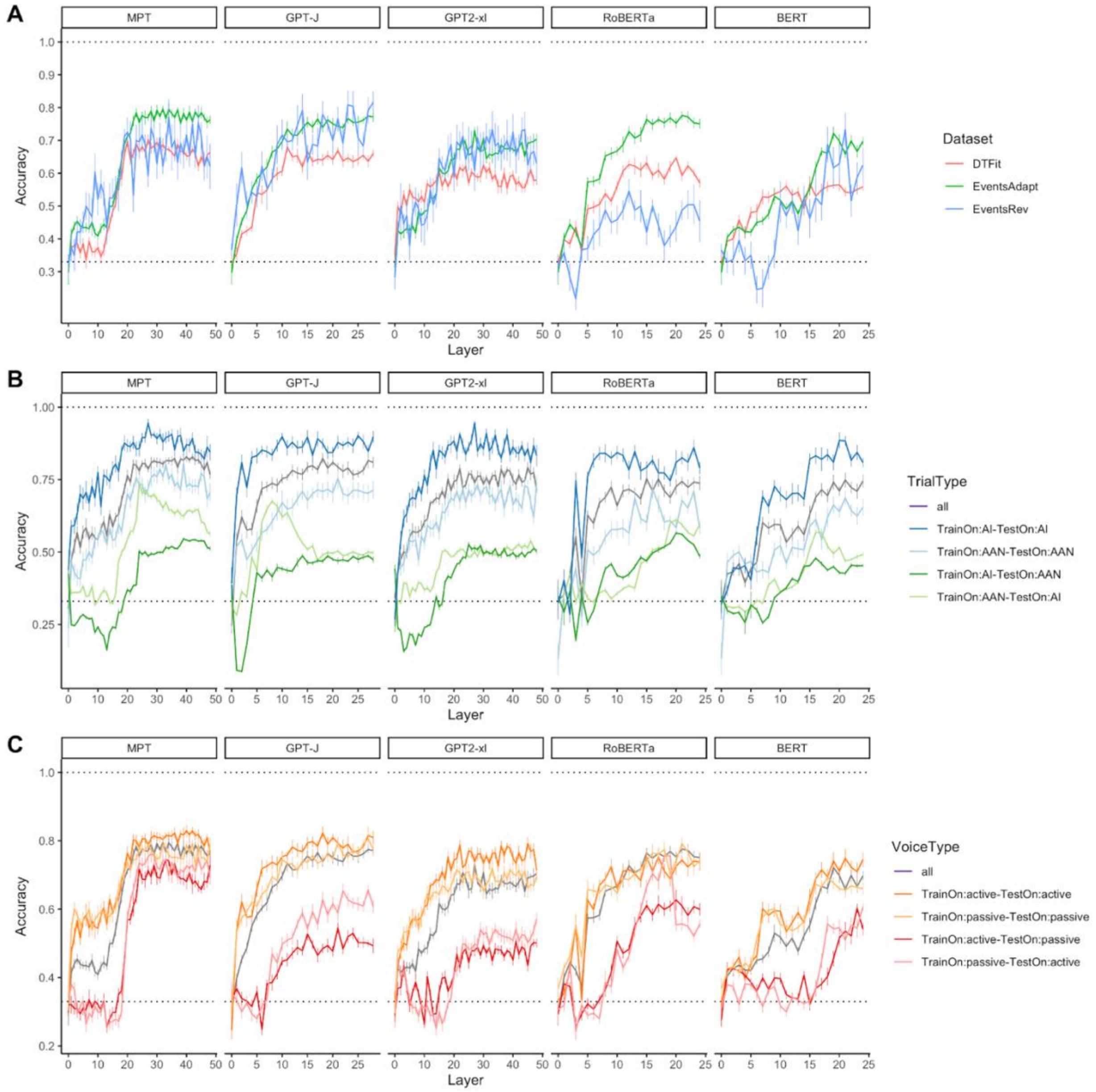

***Figure S14***. *Multiclass probing results.* ***(A)*** *Overall performance on the 3 datasets,* ***(B)*** *Generalization across sentence types, Dataset 1,* ***(C)*** *Generalization between active and passive voice sentences, Dataset 1.*